\documentclass{article}

\usepackage[preprint]{neurips_2024}

\usepackage[utf8]{inputenc} 
\usepackage[T1]{fontenc}    
\usepackage{hyperref}       
\usepackage{url}            
\usepackage{booktabs}       
\usepackage{amsfonts}       
\usepackage{nicefrac}       
\usepackage{microtype}      
\usepackage{xcolor}         
\usepackage[pdftex]{graphicx} 
\usepackage{amsmath} 
\usepackage{algorithm}
\usepackage{algpseudocode}
\usepackage{caption}
\usepackage{amsthm}
\usepackage{multirow}
\usepackage{colortbl}  
\usepackage{lipsum} 
\usepackage{wrapfig}
\usepackage{tcolorbox}
\usepackage{xspace}
\usepackage{wrapfig}
\theoremstyle{plain}
\usepackage{subcaption}
\usepackage{tabularx}
\usepackage{bbm}

\theoremstyle{definition}

\theoremstyle{remark}

\DeclareMathOperator*{\argmax}{arg\,max}

\newcommand{\fixgang}[1]{\footnote{\textcolor{red}{\textbf{FIX-GANG!!!} #1}}}
\newcommand{\fixchen}[1]{\footnote{\textcolor{red}{\textbf{FIX-CHENG!!!} #1}}}
\newcommand{\fixme}[1]{\footnote{\textbf{FIXME!!!} #1}}
\newcommand{\OurMODEL}{\textsc{Rule-Ke}}

\newcommand{\OurDATA}{\textsc{RKe-eval}}
\newcommand{\BaselineMODEL}{\textsc{Rule-Ke} (\emph{w/o} rules)}
\newcommand{\OurMODELB}{\textsc{Rb-ReasonKe}}
\newcommand{\OurMODELF}{\textsc{Rf-ReasonKe}}
\newcommand{\eat}[1]{}
\newcommand{\warn}[1]{\textcolor{red}{#1}}

\newcommand{\di}[1]{\textcolor{green}{#1}}

\newcommand{\ie}{\emph{i.e.,}\xspace}
\newcommand{\eg}{\emph{e.g.,}\xspace}

\newcommand{\etc}{\emph{etc}\xspace}

\title{Leveraging Logical Rules in Knowledge Editing: A Cherry on the Top}

\author{
  Keyuan Cheng\thanks{The first three authors contributed equally to this work.}$^{*,1,2,3}$, Muhammad Asif Ali$^{*,1,2}$, Shu Yang$^{*,1,2}$, Gang Lin$^{1,2,3}$,\\\textbf{Yuxuan Zhai}$^{1,2,3}$, \textbf{Haoyang Fei$^{1,2,3}$, Ke Xu$^{3}$, 
  Lu Yu$^{4}$, Lijie Hu\thanks{Correspondence to Lijie Hu \{lijie.hu@kaust.edu.sa\} and Di Wang \{di.wang@kaust.edu.sa\}.}$^{\dagger,1,2} $, and Di Wang$^{\dagger,1,2}$}\\
  $^1$Provable Responsible AI and Data Analytics (PRADA) Lab\\
  $^2$King Abdullah University of Science and Technology \\
  $^3$South China University of Technology \quad $^4$Ant Group 
}

\begin{document}

\maketitle

\begin{abstract}
Multi-hop Question Answering (MQA) under knowledge editing (KE) is a key challenge in Large Language Models (LLMs). While best-performing solutions in this domain use a plan and solve paradigm to split a question into sub-questions followed by response generation, we claim that this approach is sub-optimal as it fails for \emph{hard to decompose} questions, and it does not explicitly cater to \emph{correlated knowledge updates} resulting as a consequence of knowledge edits. This has a detrimental impact on the overall consistency of the updated knowledge. To address these issues, in this paper, we propose a novel framework named~\OurMODEL{}, \ie \underline{\textbf{\textsc{Rule}}} based \underline{\textbf{K}}nowledge \underline{\textbf{E}}diting, which is a cherry on the top for augmenting the performance of all existing MQA 
methods under KE. Specifically, \OurMODEL{} leverages rule discovery to discover a set of logical rules. Then, it uses these discovered rules to update knowledge about facts highly correlated with the edit. Experimental evaluation using existing and newly curated datasets (\ie \OurDATA{}) shows that \OurMODEL{} helps augment both performances of parameter-based and memory-based solutions up to 92\% and 112.9\%, respectively.
\end{abstract}

\section{Introduction}
\label{sec:intro}
Large Language Models (LLMs) have demonstrated powerful 
reasoning and comprehension capabilities for worldly 
knowledge~\cite{Huang2022TowardsRI,Wei2022ChainOT}. 
\eat{These capabilities enable LLMs to answer questions easily 
requiring multiple reasoning steps and extensive open-domain 
knowledge~\cite{Lan2021ASO,Mavi2022ASO}.}However, it has been shown that LLMs have 
a very limited adaptability to newly emerging information and/or 
knowledge~\cite{jeong2023study}.\eat{Owing to incorrect and/or outdated knowledge,}
This causes LLMs to 
generate plausible yet incorrect answers for unknown facts, 
a phenomenon known as hallucinations~\cite{Hong2023FaithfulQA,2023survey}.
Moreover, such a weakness significantly undermines the reliability of LLMs for Multi-hop 
Question Answering (MQA), which requires multiple reasoning steps and 
extensive open-domain knowledge~\cite{Lan2021ASO,Mavi2022ASO}.
\eat{However, the end-performance of LLMs is still limited by }
Updating LLMs' information/knowledge through model re-training is a 
computationally demanding tasks, making well-timed edits almost impossible. Thus, to avoid re-training, 
MQA under knowledge editing (KE), \ie answering multi-hop questions 
based on given fact edits, has thus received much attention in recent years~\cite{gu2023pokemqa,zhong2023mquake}.

Briefly speaking, there are two dominant research directions for this topic, \ie parameter-based and memory-based knowledge editing. 
Parameter-based methods update the knowledge in LLMs by modifying 
the parameters of the model. Some examples in this regard include 
ROME~\cite{meng2022locating} and its improved variant, \ie MEMIT~\cite{meng2022mass}.
On the contrary, the memory-based methods explicitly maintain 
an edit memory to store the information about the knowledge and 
facts to be modified. Examples include: MeLLo~\cite{zhong2023mquake}, 
PokeMQA \cite{gu2023pokemqa} and \textsc{Temple-MQA}\cite{cheng2024multi}. 
These methods primarily adopt plan-and-solve 
paradigm~\cite{Khot2022DecomposedPA, Wang2023PlanandSolvePI}, 
where LLMs are prompted to decompose 
a multi-hop question into multiple sub-questions followed 
by iteratively interacting with the edit memory to solve 
each sub-question.

\begin{figure}
    \centering
    \includegraphics[width=0.80\linewidth]{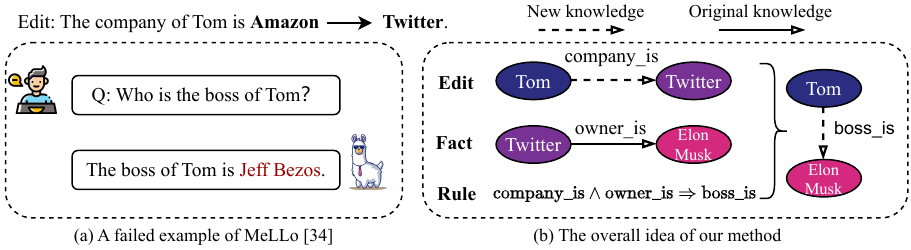}
    \caption{(a) The former company of Tom is Amazon, after updating the Tom's company to Twitter 
    the model still has past knowledge about Tom's boss. 
    (b)~\OurMODEL{} leverages logic rules to also update
    knowledge about Tom's boss to Elon Musk following the 
    edit about Tom's company.}
    \label{fig:idea}
\end{figure}

While recently plenty of work has contributed to MQA 
under KE, we observe that existing research still poses 
following key limitations: (i) These methods will fail when 
encountering "\emph{hard to decompose}" multi-hop questions, and 
(ii) they are unable to cater to "\emph{correlated knowledge 
updates}" resulting as a consequence of knowledge edits.
Here "\emph{hard to decompose}" questions imply
a subset of multi-hop questions that cannot be easily 
decomposed into independent 
sub-questions using existing plan-and-solve approaches 
(\eg \cite{zhong2023mquake,gu2023pokemqa}).
Likewise, "\emph{correlated knowledge update}" implies 
cases where we need to cater to successive knowledge updates 
resulting as a consequence of individual knowledge edits 
to come up with the correct answer.
This is also shown in Figure~\ref{fig:idea} (a), 
illustrating that it is hard for the existing 
solutions (\ie MeLLo~\cite{zhong2023mquake}) to decompose the 
the question "\emph{Who is the boss of Tom}" into 
appropriate sub-questions.
At the same time, given the fact edit to update the company of Tom from 
\emph{Amazon $\rightarrow$ Twitter}, yet after updating 
Tom's company, the language model still 
possesses outdated knowledge about the facts correlated 
with the edit, \eg Tom's current boss \etc.
This leads to knowledge inconsistency and has a detrimental 
impact on the reliability of KE methods.

To address these limitations, in this work, we propose a novel 
method, \ie~\OurMODEL{}: \underline{\textbf{\textsc{Rule}}} based \underline{\textbf{K}}nowledge \underline{\textbf{E}}diting.
As shown in Figure~\ref{fig:idea} (b),~\OurMODEL{} 
leverages logic rules to identify the core parts of the 
\emph{hard question} and performs~\emph{correlated knowledge updates},
\ie updating knowledge/facts correlated with the edit 
to ensure knowledge consistency for KE, thus correctly 
answering the question as: \emph{Elon Musk}.
To the best of our knowledge, \OurMODEL{} is amongst the 
initial attempts to employ logic rules to help augment 
the knowledge consistency and improve the end-performance 
of existing solutions for MQA under KE (both parameter-based 
and memory-based).

For experimentation, we use an existing benchmark dataset, 
as well as a newly curated dataset 
(Section~\ref{sec:benchmark}) encompassing a relatively
higher proportion of \emph{hard to decompose} questions 
and edits requiring significantly higher 
\emph{correlated knowledge updates}. Experimental 
evaluation shows that~\OurMODEL{} improves the end 
performance of existing methods by a significant margin. 

To summarize, this paper makes the following key contributions:
\begin{enumerate}
\itemsep0em
    \item We introduce~\OurMODEL{}, which leverages rule discovery 
    to discover logic rules that are helpful for \emph{hard to decompose} 
    multi-hop questions, as well as performing \emph{correlated knowledge updates} to update knowledge about facts correlated
    with the edits. 
    As a cherry on top,~\OurMODEL{} could augment the performance of all existing memory-based and parameter-based KE methods. 
    
    \item To better show the performance of \OurMODEL{}, we introduce an evaluation benchmark, \ie ~\OurDATA{}
    that, in contrast to existing benchmarks, encompasses a wide 
    range of \emph{hard to decompose} multi-hop questions and edits
    requiring significant \emph{correlated knowledge updates}.
    
    \item We perform an intensive and extensive experimental evaluation to showcase 
    the potential of our proposed approach, showing 
    that~\OurMODEL{} can augment the performance of 
    existing solutions for MQA under KE by up to 
    92\% and 112.9\% for parameter-based and memory-based 
    systems respectively.
\end{enumerate}

\section{Related Work}
\label{sec:RW}
\textbf{Parameter-based KE.} These  methods can be 
further classified into\eat{ three different methodologies}  
fine-tuning, locating and editing, and meta-learning.
Fine-tuning approaches fine-tune the model parameters or use an auxiliary set of parameters to augment the model's knowledge. 
These methods perform poorly in the KE problems because 
of vulnerability to catastrophic 
forgetting~\cite{Chen2020RecallAL, Zhu2020ModifyingMI}.
The locate and edit approaches consider the layers of a feed-forward 
network as the primary knowledge repositories and update 
their parameters to inject new knowledge. This approach was initially 
proposed in ROME~\cite{meng2022locating} targeting single edit, which is later 
extended to a large number of edits by MEMIT~\cite{meng2022mass}.
Huang et al.,~\cite{Huang2024SeeTU} emphasized the need for 
generalization across different contexts in the sub-questions. WILKE~\cite{Hu2024WilKEWK} uses a dynamic localization KE method to facilitate lifelong editing. Li et al.,~\cite{Li2023PMETPM} proposed PMET to precisely update the weights of the feed-forward network.
Meta-learning approaches treat the editing task as a 
machine learning problem.
Some examples include: hyper-network trained with 
constrained optimization for fact-only modification
\cite{DeCao2021EditingFK}, 
context-aware meta-learned loss scaling by Hu et al.,
\cite{Hu2023MetaLearningOA}, belief-graph by Hase et al.,\cite{Hase2021DoLM}. The end goal of the parameter-editing 
approaches is to develop an updated model by integrating 
information about new data/knowledge. 

These \eat{are more suitable for single-hop 
setting and} models perform poorly for multi-hop question-answering,
because 
each multi-hop question may invoke a series of inter-dependent sub-questions 
(with corresponding fact edits), which makes it harder for the parameter-based 
approaches to look for updated responses for individual sub-questions 
followed by coordinated reasoning required to generate the final answer.

\textbf{Memory-based KE.} These techniques store edits 
in an explicit memory and use retrieval-augmented methods to 
retrieve the subset of edits relevant to the question. 
For instance, Mitchell et al.,~\cite{Mitchell2022MemoryBasedME} 
proposed SERAC, a semi-parametric editing method coupled with 
retrieval-augmented counter-factual model. 
Cheng et al.,~\cite{Zheng2023CanWE} employed in-context 
learning based on three types of demonstrations (copy, update, 
and retain) to edit the model’s knowledge. 
Han et al.,~\cite{han-etal-2023-improving} developed an editing 
retrieval framework to promote the efficacy of model editing methods 
during sequential edits.
Zhong et al.,~\cite{zhong2023mquake} proposed MeLLo that uses 
plan-and-solve paradigm along with self-checking\eat{, i.e., they 
use language model} to check whether the retrieved edit is 
relevant to the question.
Gu et al.,~\cite{gu2023pokemqa} proposed PokeMQA, a two-stage retrieval by 
decoupling sub-question decomposition from knowledge editing. TEMPLE-MQA~\cite{cheng2024multi} transforms the multi-hop questions and edits into a structured form, effectively improving the accuracy of retrieval of relevant edits while addressing the challenge of temporal knowledge editing.

These methods primarily rely on the plan-and-solve paradigm, \ie 
using existing LLMs to decompose the question into sub-questions, 
later, use an inference plan for response/answer generation. They perform poorly on \emph{hard to decompose} 
multi-hop questions, which limits their ability to 
ensure knowledge consistency followed by knowledge edits.
This has a detrimental impact on the end-performance of 
MQA under KE systems.

\textbf{Logical Rules for Knowledge Graph Reasoning.} Logical 
rules have extensively been studied for the knowledge 
graph reasoning tasks.
Quinlan et al.,~\cite{Quinlan1990LearningLD} used inductive logic 
programming to derive logical rules from training samples in the 
knowledge graph. 
RNNLogic~\cite{Qu2020RNNLogicLL} trains rule generator and reasoning 
predictor to generate high-quality logic rules. 
While most of the above-mentioned works focus on mining rules 
from the knowledge graph triples, RuLE by Tang et al.,~\cite{Tang2022RulENK} 
focuses on the setting where rules are provided beforehand and 
leverages these provided rules to augment the inference performance 
of the model. Lajus et al.,~\cite{Lajus2020FastAE} proposed a rule mining system AMIE-3, which employs a range of pruning strategies and optimization methods to enable fast and accurate rule mining. So far, we are not aware of any work using logic rules 
for knowledge editing systems.

To summarize, we argue that in contrast to the existing work, 
\OurMODEL{} leverages rule discovery to infer logic rules 
that are helpful in inferring new knowledge 
based on the correlation between the given edit and the model's 
prior knowledge.
This makes~\OurMODEL{} a better choice to ensure knowledge 
consistency, which is helpful in augmenting the performance of existing solutions 
for MQA under KE 
for both parameter-based and memory-based settings.

\eat{Rule research in the field of knowledge graph reasoning is relatively complete, but previous research has ignored its importance to knowledge editing. \OurMODEL{} use the mature techniques for mining rules and generate other knowledge influenced by editing.}

\eat{Specifically, for parameter-based methods we can leverage
the inferred knowledge to further edit the model's 
knowledge/information. Whereas, for memory-based methods,
we can store the inferred facts in the edit memory, thus 
allowing the model to generate correct response
for complex inter-related knowledge thus improving the end-performance of the model.}

\eat{\OurMODEL{} can also be used as a plug-and-play module to improve the robustness of memory-based methods to plans. By storing the inferred coarse-grained knowledge in the edit memory, the model is able to retrieve the corresponding knowledge despite the coarse-grained plan, which improves the accuracy of answering multi-hop questions}

\eat{The parameter-based method can leverage these new knowledge to edit the model, thereby making these methods suitable for multi-hop questions.}
\section{Preliminaries}
\label{sec:preliminaries}
In this section, we introduce the notations, 
provide the motivation to use logical rules 
for~\OurMODEL{}, and define our problem. 

{\bf Notations.}
We use $\mathcal{K}_{base} = \{r(s,o) | s,o \in \mathcal{E}, 
r \in \mathcal{R}\}$
to represent the knowledge base, where $s$ represents the 
subject entity and $o$ represent the object entity of 
the relation $r$, $\mathcal{E}$ represents the set of 
entities and $\mathcal{R}$ represents set of relations. 
We use $e=r(s,o \rightarrow o^*)$ to represent an individual 
knowledge edit, denoting that object of the subject $s$ with 
relation $r$ is updated from $o$ to $o^*$. A collection of  $n$
edits is represented by $\mathcal{M}=\{e_1,e_2,\cdots, e_n\}$. 
We use $f$ to represent a language model and use $Q$ to 
represent a multi-hop question. Each $Q$ requires multiple 
reasoning steps to obtain the final answer, where the reasoning 
steps form a \textit{knowledge path} $\mathcal{P}_ Q=\langle 
r_1(s_1,o_1),\cdots, r_n(s_n,o_n)\rangle$ with $o_n$ as the 
final answer. If one of the fact $r_i(s_i,o_i) \in \mathcal{P}_Q$ is updated by 
an edit $e_i = r_i(s_i, o_i \rightarrow o_i^*) \in \mathcal{M}$, 
the resulting knowledge path becomes:
$\mathcal{P}^{*}_Q = \langle r_1(s_1,o_1),\cdots, r_i(s_i, o_i^*),\cdots,r_n(s_n^*, o_n^*)\rangle$, \ie the subsequent knowledge path will be changed, 
yielding $o_{n}^{*}$ as the new answer for question $Q$. It's notable that $Q$ may be modified by multiple edits in $\mathcal{M}$.

\eat{However, previous work overlooks that one edit not only may affect the reasoning process of $Q$ but also may introduce a new knowledge edit, i.e., an edit $e=r(s,o\rightarrow o^*)$ may lead to a new edit $e^\prime$. For example, $e=\text{Company\_is}(\text{Tom},\text{Amazon}\rightarrow \text{Twitter})$  can lead to $e^\prime=\text{Boss\_is}(\text{Tom},\text{Jeff Bezos}\rightarrow\text{Elon Musk})$. This is because knowledge is not an individual unit, and they are related to each other. Thus, in this paper, we aim to leverage logical rules to represent their association explicitly.} 

We use $\varphi: X \rightarrow {p_0}$ to represent a logic rule, 
where $X$ is a conjunction of predicates representing 
{\em pre-condition}, and $p_0$ is a predicate representing 
{\em consequence}. 
In this work, we use compositional logic rules of the form: 
\begin{equation}
    \varphi(r_h, \textbf{r}_\textbf{b}): r_{b_{1}}(z_0,z_1) \land r_{b_{2}}(z_1,z_2) \land \cdots \land r_{b_{n}}(z_{n-1},z_n) \rightarrow r_h(z_0,z_n)
\end{equation} 
where $\textbf{r}_\textbf{b}=\{r_{b_1},\cdots,r_{b_n}\}$ is a set 
of relations for {\em pre-condition}, and $r_h$ is a relation for 
{\em consequence}, $\{z_0,\cdots,z_n\}$ is a set of variables. 
When these variables are replaced with specific entities, we can 
get a \textit{grounding} of rules. For example: 
\emph{{father$\_$is}({Tom},{John}) $\wedge$ {wife$\_$is}({John},{Amy}) $\rightarrow$ {mother$\_$is}({Tom},{Amy})}. This 
grounding illustrates that if Tom's father is John and 
John's wife is Amy, then it implies Tom's mother is Amy. 

\noindent{\bf Motivation of Logic Rules.}
The motivation for employing logic rules for \OurMODEL{}
stems from the fact that
an individual edit $e=r(s,o\rightarrow o^*)$ not only 
affects the knowledge path of $Q$, but may also initiate 
new knowledge edits $e^\prime$.
For instance, an edit
\emph{$e$={company$\_$is}({Tom},{Amazon} $\rightarrow$ {Twitter})}
will also trigger
\emph{$e^{'}$={boss$\_$is}({Tom},{Jeff Bezos} $\rightarrow$ {Elon Musk})}.
For such cases, logic rules provide an effective 
mechanism to capture and represent these associations
among edits more explicitly, which makes them an ideal 
choice for~\OurMODEL{}.
We call this phenomenon as 
\emph{correlated knowledge updates} for edit $e$
in the rest of this paper.
Note that this issue has been overlooked by previous 
research on KE.

\textbf{Problem Definition.}
MQA under KE aims to inject new knowledge 
$\mathcal{M}$ into the language model $f$ to come up with
an updated model $f^*$ to be used to answer multi-hop questions $Q$.
For this, we first use rule discovery tools to discover the logic 
rules $\Sigma=\{\varphi_1,\varphi_2, \cdots\,\varphi_m \}$ 
from existing knowledge-base $\mathcal{K}_{base}$. 
Later, we use these rules $\Sigma$ in relation with the 
edit information $\mathcal{M}$ to discover the knowledge updates 
correlated with the edits as augmented knowledge $\mathcal{K}^{'}$. 
Finally, we use the $\mathcal{K}^{'}$ along with $\mathcal{M}$
to perform knowledge editing for existing methods in a 
performance-enhanced fashion.

\section{\OurMODEL{}: \underline{Rule} Based \underline{K}nowldege 
\underline{E}diting}
\label{sec:proposed}

The core idea of~\OurMODEL{} is to leverage logical rules 
to infer new/updated knowledge, resulting in a correlation between requested knowledge edits and prior knowledge.\eat{Note 
that~\OurMODEL{} is a plug-and-play framework that could be 
employed to augment the performance of existing knowledge 
editing methods and maintain knowledge consistency.} Its 
workflow is explained as follows.

\subsection{Workflow of~\OurMODEL{}}
The overall workflow of~\OurMODEL{} is shown in 
Figure~\ref{fig:framework}. 
It could be explained as a three-step process:
(a) Discover logic rules $\Sigma$ from a large-scale 
knowledge base $\mathcal{K}_{base}$.
(b) For each edit $e \in \mathcal{M}$, 
\eat{For the logic rules $\Sigma$,} 
determine the subset of rules correlated by the 
edit $e$, \ie $\Sigma_{e} \subseteq \Sigma$.   
(c) Use correlated rules $(\Sigma_{e})$ to infer 
and/or correlated knowledge to be stored as augmented knowledge $\mathcal{K}^{'}$, later used for knowledge consistency of
KE methods.

\textbf{Step a: Mining Logic Rules.} 
The first step of~\OurMODEL{} is the rule mining process. 
It aims at excavating the common logic rules $\Sigma$ from 
the large-scale knowledge base $\mathcal{K}_{base}$, 
facilitating subsequent prediction of the 
\emph{correlated knowledge} affected by the edits. 
For this, we use Wikidata~\cite{2014wikidata} as our 
knowledge base. The quality of the discovered rules 
varies and is not always valid in all situations.
In order to avoid redundant rules, 
we use \textit{support threshold} 
$\mathcal{A}_{\Sigma}$ as a quantitative metric, 
indicative of the quality of the rule.

There are many works that study rule mining and achieve excellent results, such as AMIE-3~\cite{Lajus2020FastAE} and RNNLogic~\cite{Qu2020RNNLogicLL}. Thus, we use the existing tools to mine logic rules~\eat{$\Sigma$ } and their corresponding \textit{support threshold} from $\mathcal{K}_{base}$, as follows:
\begin{equation}
  \Sigma, \mathcal{A}_{\Sigma}= \text{Tool$_{\text{mine}}$}(\mathcal{K}_{base}),
\end{equation}
where $\Sigma=\{\varphi_1,\varphi_2, \cdots\,\varphi_m \}$ is a set of rules, $\mathcal{A}_{\Sigma}=\{\alpha_{\varphi_1}, \alpha_{\varphi_2},\cdots, \alpha_{\varphi_m}\}$ is a set of support thresholds, with $\alpha_{\varphi_i}\in [0, 1]$ indicating the likelihood of the rule $\varphi_i$ to be true. 
We use AMIE-3 as our rule mining tool $(\text{Tool$_{\text{mine}}$})$, 
as it has the ability to mine higher-quality rules in practice. 

\textbf{Step b: Determining correlated Rules.}
Editing not only affects single knowledge but may trigger 
changes in the entire knowledge system, causing its related 
knowledge to be updated accordingly. The logic rules serve 
as a bridge between an edit and its \textit{correlated knowledge}, 
thus, determining the relevant bridge for each edit is very 
important. The objective of this step is to find a subset of 
relevant rules $\Sigma_{e} \subseteq \Sigma$ for each edit 
$e \in \mathcal{M}$. To determine the subset of rules 
that are relevant to an edit, we use the semantic similarity 
between the relation part of the edit and the relations in 
the pre-conditional part of the rule as a selection criterion.
\eat{We use a set of rules 
$\varphi_{i}{(r_h,\mathbf{r_b}) \in \Sigma}$ to demonstrate 
the relevance of the rule $\varphi_{i}$ and edit $e$.}
For instance, given an edit 
"\emph{work$\_$for (Tom, Apple) $\rightarrow$ Twitter}",
and a set of rules: 
\{"\emph{company$\_$is ($z_0$,$z_1$) $\land$ owner$\_$is 
($z_1$,$z_2$) $\rightarrow$ boss$\_$is ($z_0$,$z_2$)}";
"\emph{born$\_$in ($z_0$,$z_1$) $\land$ locate$\_$in 
($z_1$,$z_2$) $\rightarrow$ raised$\_$in ($z_0$,$z_2$)}"\},
the relation part of the edit "\emph{work$\_$for}" 
has higher semantic relevance with the relation
"\emph{company$\_$is}" compared to other relations, 
\eg "owner\_is", "born\_in" \etc.

To implement our above intuition, we followed the idea of dense retrieval~\cite{Karpukhin2020DensePR} to contrast the semantic 
relevance based on the vector similarity of the relation embeddings. 
For this, we use an encoder $E$ to encode the relation $r$ of the 
edit $e$ and the relation set $\mathbf{r_b}$ of rule 
$\varphi_{}{(r_h,\mathbf{r_b})}$, as 
shown below:
\begin{equation}
    \begin{aligned}
        v_{e} & = E(r), \\
        \mathbf{V}_{\varphi_{}} & = E(\mathbf{r_b}) = \{ E(r_{b_1}),\cdots ,E(r_{b_n})\},
    \end{aligned}
\end{equation}

\eat{
\begin{equation}
  v_{r_{i}}=E(r_{i}),
\end{equation}
\begin{equation}
 \mathbf{V}_{\mathbf{r_b}} = \{ E(r_{b_1}),\cdots ,E(r_{b_n})\}, 
\end{equation}}

where $v_e$ is the vector embedding for relation $r$ of edit $e$, $\mathbf{V}_{\varphi_{}}$ is the set of vector 
embedding for the relation set $\mathbf{r_b}$ of $\varphi_{}$. Later, we use cosine similarity ($\text{sim}$) for the elements of 
$\mathbf{V}_{\varphi_{}}$ and $v_e$ to determine the relational predicate exhibiting maximum similarity with relation $r$, as shown below. 
\begin{equation}
    v^{*}_k = \argmax_{v_k \in \mathbf{V}_{\varphi_{}}} {(\text{sim}(v_k, v_e))},
\end{equation}
where $v^{*}_k$ is the embedding vector of the 
predicate relation $r_{b_{k}}$ with highest similarity 
with $v_e$. 

Because the quality of rule $\varphi_{}$ may not be high and \textit{correlated knowledge} found by low-quality rules 
is not necessarily reliable, we \eat{should }use $\alpha_\varphi$ 
times scaled similarity of 
$v^{*}_k$ and $v_e$ compared against the 
correlation threshold
$\delta$ to determine whether $\varphi_{i}(r_h,\mathbf{r_b})$ is a reliable rule for predicting \textit{correlated knowledge} for subsequent steps: 
\begin{figure}
    \centering
    \includegraphics[width=0.80\linewidth]{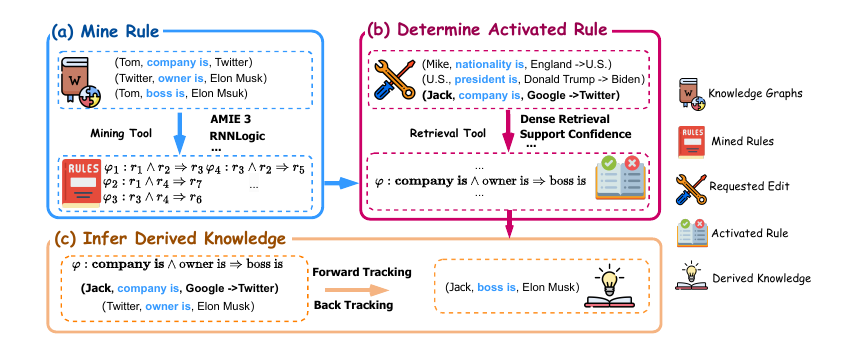}
    \caption{The overall workflow of~\OurMODEL{}. It encompasses: 
    (a) rule mining, (b) determining correlated rules, and 
    (c) inferring correlated  knowledge.} 
    \label{fig:framework}
\end{figure}
\begin{equation}
  \psi: (v^{*}_k \cdot v_e) * \alpha_{\varphi} > \delta,
  \label{Eq:threshold}
\end{equation}
where $\psi$ returns a logical indicator of whether the 
the statement is true or not. Finally, we use the above process to find all the logic rules $\varphi_{}{(r_h,\mathbf{r_b}) \in \Sigma}$ which satisfy corresponding $\psi$ is true, and combine them into the rule subset $\Sigma_{e}$, \ie a set of rules activated by edit $e$. 


\textbf{Step c: Inferring Correlated Knowledge.} 
After selecting the subset of rules 
$\Sigma_{e}$ corresponding to the relation in the edit $e \in \mathcal{M}$, \eat{now }we need to infer \emph{correlated knowledge} for edit $e$.  
\eat{This step aims to obtain correlated knowledge-based rule 
$\varphi(r_h, \textbf{r}_\textbf{b})$ and its correlated 
edit $e = r(s,o\rightarrow o^*)$.}
Suppose the rule is activated at position $k$, \ie 
at pre-conditional predicate $r_{b_{k}}$\eat{.  Then}, we can instantiate variables $z_{k-1}$ and $z_k$ as $s$ and $o^*$, respectively, as shown below. 
\begin{equation}
    r_{b_{1}}(\color{teal}\mathbf{z_0}\color{black},z_1) \land r_{b_{2}}(z_1,z_2)\cdots \underbrace{r_{b_k}(z_{k-1}(s),z_k(o^*))}_{\textsc{New Knowledge}}  \cdots\land r_{b_{n}}(z_{n-1},\color{brown}\mathbf{z_n}\color{black}) \rightarrow \underbrace{r_h(\color{teal}\mathbf{z_0}\color{black},\color{brown}\mathbf{z_n}\color{black})}_{\textsc{correlated Knowledge}}.
\end{equation} 
The above equation shows how the \emph{correlated knowledge} is 
generated from the rule $\varphi$ and edit $e$. However, we only know $z_{k-1}=s$ and $z_k=o^*$. Thus, we need to find the specific entity for variable $\color{teal}\mathbf{z_0}\color{black}$ and $\color{brown}\mathbf{z_n}\color{black}$ to determine $r_h(\color{teal}\mathbf{z_0}\color{black},\color{brown}\mathbf{z_n}\color{black})$ as the \emph{correlated knowledge}. For this, we use:
(i) back tracking to determine the $\color{teal}\mathbf{z_0}\color{black}$; and  
(ii) forward tracking to determine the $\color{brown}\mathbf{z_n}\color{black}$, with  
corresponding chain shown in \eqref{Eq:bwfw_chain}.
\begin{equation}
\underbrace{\color{teal}\mathbf{z_0}\color{black},z_1,\cdots,z_{k-1}(s)}_{\textsc{Back Tracking}\leftarrow}, \underbrace{z_{k}(o^*),\cdots,z_{n-1},\color{brown}\mathbf{z_n}\color{black}}_{\rightarrow \textsc{Forward Tracking}}.
\label{Eq:bwfw_chain}
\end{equation}
The process flows of forward tracking and backward tracking are 
explained as follows. 

\textbf{Forward Tracking} is an iterative process used to determine the specific entity for the last variable $\color{brown}\mathbf{z_n}\color{black}$ by traversing the chain $(z_k(o^*),\cdots,z_{n-1},\color{brown}\mathbf{z_n}\color{black})$ 
in the forward direction for one step at a time, with only prior knowledge 
about the specific entity $o^*$ for $z_k$. 
Each step of this iterative process assumes that we aim to
determine the specific entity of variable $z_{m+1}$ based on $r_{b_{m+1}}(z_m(o_{z_m}),z_{m+1})$, where $o_{z_m}$ is the specific entity for $z_m$, and $m$ indicates
the current iterative position $k \leq m \leq n-1$. 

It is easy to find that each step corresponds to a single-hop question. However, this question may be modified by the existing edits. Thus, we need to query the edit memory if there are related edits based on $o_{z_m}$ and $r_{b_{m+1}}$.
For this, we use dense retrieval \cite{Karpukhin2020DensePR} to 
determine the corresponding edit, with details provided in 
Appendix~\ref{Appendix:retrieval}.

For the cases, we are unable to find the relevant edit in edit memory, 
we will query a language model to infer the specific entity for variable $z_{m+1}$. For this, we design 
an in-context learning prompt (explained in Appendix~\ref{sec:prompt}) 
that prompts an LLM as follows.
\begin{equation}
 o_{z_{m+1}} = \text{LLM}(\text{P}_{\text{forward}}(o_{z_m},r_{b_{m+1}})),
 \label{eq:forward}
\end{equation}
where $\text{P}_{\text{forward}}$ is the in-context prompt for $o_{z_m}$ and 
relation $r_{b_{m+1}}$ used as input for LLM, yielding $o_{z_{m+1}}$ as specific entity for variable $z_{m+1}$. We can finally get the specific entity for $\color{brown}\mathbf{z_n}\color{black}$ by iterating the above process. 

\textbf{Back Tracking} is similar to forward tracking. The primary difference 
is that it traverses the chain $(\color{teal}\mathbf{z_0}\color{black},\cdots,z_{k-1}(s))$ in the reverse direction
in order to infer the specific entity for $\color{teal}\mathbf{z_0}$. Specifically, 
each step of this process assumes that we aim to
determine the specific entity of variable $z_{j-1}$ based on $r_{b_{j}}(z_{j-
1},z_j(o_{z_{j}}))$, where $o_{z_j}$ is the specific entity for $z_j$, and $j$ indicating the current iterative position and $1 \leq j \leq k-1$.

Here again, similar to the forward tracking, we first look for relevant
edits in the edit memory helpful in inferring $z_{j-1}$ based on 
$o_{z_j}$ and $r_{b_{j}}$.
\eat{there are also two situations for the  of $z_{j-1}$.}
This is explained in detail in Appendix \ref{Appendix:retrieval}.
For the cases where we are unable to find relevant edits, we 
infer $z_{j-1}$ by querying the language model.
For this, we construct an in-context learning prompt that prompts 
LLM to infer $z_{j-1}$:

\begin{equation}
 o_{z_{j-1}} = \text{LLM}(\text{P}_{\text{back}}(r_{b_{j}},o_{z_j}))
 \label{eq:back}
\end{equation}

where $\text{P}_\text{back}$ is the in-context prompt 
that uses $r_{b_{j}}$ and $z_j$ as input and prompts 
LLM, to yield $o_{z_{j-1}}$ as specifc entity for $z_{j-1}$, for details refer 
to Appendix \ref{sec:prompt}. 
\eat{Finally, we will get the specific entity for $\color{teal}\mathbf{z_0}\color{black}$.}

Finally, we use the start point $\color{teal}\mathbf{z_0}\color{black}$ and endpoint $\color{brown}\mathbf{z_n}\color{black}$ resulting from 
backward tracking and forward tracking to infer the correlated knowledge 
$r_{h}(\color{teal}\mathbf{z_0}\color{black},\color{brown}\mathbf{z_n}\color{black})$ to be inserted in $\mathcal{K}^{'}$.

\subsection{\OurMODEL{}---\emph{A cherry on the Top}}
Existing KE methods can not maintain knowledge consistency effectively. The augmented knowledge $\mathcal{K}^{'}$ can 
help to alleviate this problem because it is generated 
from the interactions between edits $\mathcal{M}$ and 
logic rules $\Sigma$. 
In order to adapt~\OurMODEL{} to downstream KE methods, we 
need to transform the knowledge in $\mathcal{K}^{'}$ from 
knowledge to edits, \ie from 
$r(s,o)$ to $r(s,o\rightarrow o^*)$. Specifically, we define a new set of edits based on $\mathcal{K}^{'}$ as follows: 
\begin{equation}
 \mathcal{M}_{\mathcal{K}^{'}} =\{r(s,\text{null}\rightarrow o) \mid r(s,o) \in \mathcal{K}^{'}\}, 
 \label{eq:transform}
\end{equation}
where $\text{null}$ is a specific placeholder, meaning that we no 
longer need the information about the original object.
We then merge the newly curated edit set 
$\mathcal{M}_{\mathcal{K}^{'}}$ with $\mathcal{M}$ to
come up with $\mathcal{M}_{\text{AUG}}$, \ie $\mathcal{M}_{\text{AUG}}=\mathcal{M}\cup \mathcal{M}_{\mathcal{K}^{'}}$ as the final set 
of knowledge edits.
Finally, we leverage $\mathcal{M}_{\text{AUG}}$ to further 
edit the model's knowledge/information for parameter-based 
methods, whereas for memory-based methods, we can store $\mathcal{M}_{\text{AUG}}$ in the edit memory, thus allowing 
the model to generate the correct response for correlated 
knowledge instances.

\section{\OurDATA{}: \OurMODEL{} Evaluation benchmark}
\label{sec:benchmark}
We observe that existing benchmarks for MQA under KE, \eg~\textsc{MQuAKE}~\cite{zhong2023mquake}, 
are not an ideal option to rigorously test the ability to maintain knowledge consistency for editing methods.
The majority of the data in \textsc{MQuAKE} can 
easily be decomposed into multiple independent knowledge
sequences/units, which can be answered independently 
to come up with the final answer. 
This severely undermines the end-utility of existing 
benchmarks for analyzing: (i) hard-to-decompose questions, and 
(ii) correlation among knowledge updates, 
\ie how an individual knowledge update may 
impact the knowledge for the subsequent parts.

\setlength{\intextsep}{0pt plus 3pt}
\begin{wrapfigure}{r}{0.4\textwidth}
  \centering
  \includegraphics[width=0.4\columnwidth]{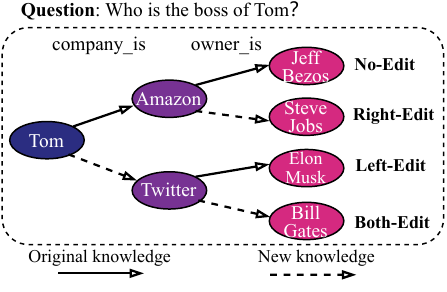}
  \vspace{-3.7ex}
  \caption{Four different scenarios of new knowledge 
   generated by logic rules and edits for \OurDATA{}.}
\label{fig:datasets}
\end{wrapfigure}

In order to better evaluate the knowledge consistency of 
\OurMODEL{} and existing methods on MQA under KE for 
"\emph{hard questions}" and "\emph{correlated knowledge updates}", we propose a new evaluation benchmark, 
namely:~\textsc{\underline{\textbf{R}}ule\underline{\textbf{Ke}}}-\textsc{\underline{\textbf{Eval}}uation} 
(\OurDATA{}).
\OurDATA{} encompasses numerous multi-hop questions and 17 
different rules with pre-condition of length 2. Each multi-hop question is associated 
with a unique rule and at least one edit. 
As shown in Figure~\ref{fig:datasets}, we categorize
the knowledge inferred in~\OurDATA{} 
into four different scenarios: 
(i) No-Edit: the knowledge of the rule part is not 
modified by edit. 
(ii) Right-Edit: only the second part of the rule knowledge 
is modified by edit and requires back tracking to answer 
this question. 
(iii) Left-Edit: only the first part of the rule knowledge 
is modified by edit and requires forward tracking to 
answer this question. 
(iv) Both-Edit: all the knowledge of the rule part is modified 
by edit.
The process flow of data curation is explained in detail in Appendix~\ref{Appendix:data}, along with an example 
illustration in Appendix Table~\ref{tab:data_example}. 
This categorization method comprehensively covers all scenarios 
where new knowledge is generated due to edits and corresponding 
logic rules.


\eat{
\OurDATA{} is explicitly designed to evaluate 
whether the model can maintain knowledge consistency 
during knowledge editing.}

\eat{Specifically, they are  (i) 
the correlation among knowledge updates, 
\ie how an individual knowledge update may also impact the 
knowledge for the subsequent parts, which is crucial in order
to ensure the knowledge consistency for the knowledge editing 
methods.}

\eat{for the end methods
This severely undermines the end-utility of these benchmarks 
to rigorously evaluate the knowledge consistency, 
especially for highly inter-related and hard questions.
knowledge which undermines the abilities of 
for knowledge consistency
for knowledge consistency.}

\eat{For majority 
When answering a multi-hop question, the model only needs to 
perform step-by-step inference step by step 
based on the model's existing knowledge or external 
editing knowledge, which is not sufficient to ensure 
the consistency of knowledge under knowledge editing.}
\section{Experiments}
\label{sec:exp}
\subsection{Experimental Settings}
\label{EXP:Seting}

\textbf{Datasets.}
We evaluate the performance of~\OurMODEL{} using
an existing knowledge editing 
benchmark~\textsc{MQuAKE}-CF-3K~\cite{zhong2023mquake} and 
our proposed dataset~\OurDATA{}. Detailed descriptions and statistics of these datasets are provided in Appendix~\ref{Appendix:dataset}.

\textbf{Evaluation Metrics.}
We use multi-hop accuracy (Acc)~\cite{zhong2023mquake} as 
an evaluation metric.
\eat{ in order to compute the accuracy of the final 
answer on multi-hop questions}
Each instance in \textsc{MQuAKE}-CF-3K include three multi-hop 
questions. An instance is considered correct if any of the three 
multi-hop questions is answered correctly. 
Mathematical formulation and further details about the 
evaluation metrics are given in Appendix~\ref{Appendix:evaluation}.
Note that just for simplification, we do not use 
hop-wise accuracy~\cite{gu2023pokemqa} as a metric because our method primarily relies on the final response rather than 
the evaluating the intermediate path.

\textbf{Baseline methods.}
For performance evaluation of~\OurMODEL{}, we use both 
parameter-based and memory-based KE methods as baselines. 
The parameter-based baselines include:
{Fine-tuning (FT)}~\cite{Zhu2020ModifyingMI},
{ROME}~\cite{meng2022locating} and
{MEMIT}~\cite{meng2022mass}.
The memory-based baselines include 
{MeLLo}~\cite{zhong2023mquake}, {PokeMQA}~\cite{gu2023pokemqa} and {TEMPLE-MQA}~\cite{cheng2024multi}. 
Details about the baseline models are provided in 
Appendix~\ref{Appendix:baseline}.

\eat{In addition, in order to better enhance the knowledge consistency of the model in the absence of rules, we propose a simple baseline method dubbed \OurMODEL{} 
(\emph{w/o} rules). 
Further details of this method are explained in 
Appendix~\ref{Appendix:model-wo-rules}.}


\textbf{Experimental Setup.} 
To assess performance across different number of edits, 
we conduct stratified sampling based on the number of hops,
which allows us to construct batches of different sizes.
The setting of different batch sizes is denoted as 
$k$-E $(k\in\{1,100,All\})$. 
The value of the correlation threshold $\delta$ is 0.8.
The value of $\theta$ in Appendix~\ref{Appendix:retrieval} is 0.7.
We conduct experiments on 
the three language models, \ie \textsc{LLaMa-2-7B}~\cite{Touvron2023Llama2O}, \textsc{LLaMa-2-7B-chat}~\cite{Touvron2023Llama2O} 
and \textsc{GPT-turbo-3.5-instruct}~\cite{Ouyang2022TrainingLM}.
Further details about these models are given in 
Appendix~\ref{Appendix:LLMs}. We evaluate all the methods 
in the $\{1,100,All\}$-E setting. All experiments are 
performed using PyTorch 2.1.0 with two A40-48GB GPUs.  
All the results reported in the paper are averaged over three runs.

\subsection{Experimental Results}
\paragraph{\OurMODEL{} can boost the performance of memory-based methods.} 

We first analyze the impact of~\OurMODEL{} in augmenting
the performance of memory-based methods.
Corresponding results in Table~\ref{tab:exp2_1} show 
that~\OurMODEL{} 
consistently augments the performance 
of the memory-based methods, outperforming the 
original models by a significant margin. 

Considering the results of our model using~\textsc{LLaMa-2-7B},
\OurMODEL{} improves the performance by up to 
\{112.92\%, 106.31\%, and 95.20\%\} for~\OurDATA{}, and 
\{12.13\%, 2.30\%, and 2.03\%\} for~\textsc{MQuAKE}
compared to the best-performing original variants 
under \{1, 100, and All\}-E settings respectively. 
The results of variants for~\OurMODEL{} using
\textsc{GPT-turbo-3.5-instruct} exhibit a similar trend. 
Such drastic improvement in performance showcases the 
logical generalization abilities of~\OurMODEL{}, ensuring 
knowledge consistency for MQA under KE.

We can also observe that as the size of the edit batch 
increases, the improvement of performance decreases 
gradually compared to smaller edit batches.
A key justification in this regard is the fact that 
logic rules leveraged by~\OurMODEL{} generate 
additional knowledge edits, thus increasing the size of the edit 
memory by more than three-fold, which makes it cumbersome for 
the retrieval to find the right edit from the edit memory.
Nevertheless, this issue could be mitigated by using robust 
retrieval systems. For instance, TEMPLE-MQA~\cite{cheng2024multi}, 
with a better retrieval performance, outperforms MeLLo and 
PokeMQA in the 100/all-E scenario on \OurDATA{}.

\eat{ \textbf{\OurMODEL{} can augment logic 
generalization for parameter-based method.}}

\begin{table*}[!tb]
    \centering
    \renewcommand{\arraystretch}{1.2} 
    \scalebox{0.68}{
    \begin{tabular}{cc|cccccc|cccccc}
    \specialrule{1pt}{0pt}{0pt} 
    \multirow{3}{*}{\textbf{AUG}} &  \multirow{3}{*}{\textbf{Method}}  & \multicolumn{6}{|c|}{\cellcolor{gray!25}\scshape LLaMa-2-7B} & \multicolumn{6}{|c}{\cellcolor{gray!25}\scshape GPT-3.5-turbo-instruct}   
    \\ \cline{3-14} 
    & &\multicolumn{3}{c}{\textbf{\textsc{MQuAKE}-CF-3K}} & \multicolumn{3}{c|}{\textbf{\OurDATA{}}} & \multicolumn{3}{|c}{\textbf{\textsc{MQuAKE}-CF-3K}} & \multicolumn{3}{c}{\textbf{\OurDATA{}}}      
    \\ \cline{3-14}

    & & 1-E    & 100-E      & All-E  & 1-E    & 100-E      & All-E & 1-E    & 100-E      & All-E  & 1-E    & 100-E      & All-E 
    \\ \hline
    \multirow{3}{2cm}{\centering Original}
    & MeLLo     & 36.13   & 28.50    & 21.33   &18.24 &13.84 &2.03     & 59.67   & 41.47    & 35.81   &29.05 &23.73 &10.14 \\
    & PoKeMQA     & 39.13   & 29.33  & 22.83   &\underline{19.81} &14.82 &3.81   &66.18 &55.89 &47.42  &32.05 &25.18 &13.81\\
    & TEMPLE-MQA    &\underline{46.67}  & \underline{43.95} & \underline{41.33} & 16.54 &\underline{15.84} & \underline{14.19}  & \underline{75.08}  &\underline{66.02}  & \underline{53.12}  & \underline{37.29} & \underline{35.97} &\underline{33.56}\\
    \hline
    \multirow{3}{2cm}{\centering\OurMODEL{}}
    & MeLLo     &51.33 &30.28 &26.13   &{41.22} &16.27 &14.26     &69.33 &52.71 &42.06   &62.51 &45.65 &32.43  \\
    & PoKeMQA     &51.17 &29.71 &27.91   &\textbf{42.18} &{19.71} &{16.83}   &73.41 &60.95 &52.38  &{63.61} &{47.41} &37.72  \\
    & TEMPLE-MQA    &\textbf{52.33} &\textbf{44.96} &\textbf{42.17}   &38.79 & \textbf{32.68} & \textbf{27.70}  &\textbf{80.73} &\textbf{69.06} &\textbf{58.03}   &\textbf{65.21} &\textbf{61.18} &\textbf{54.68} \\
    \hline
    \specialrule{1pt}{0pt}{0pt} 
    \end{tabular}}
    \caption{Results for memory-based methods, \ie original and 
    augmented with~\OurMODEL{} on the \textsc{MQuAKE}-CF-3K and \OurDATA{}.
    We use multi-hop accuracy (Acc) as the evaluation metric. The best scores are \textbf{bold-faced}, with the best results among the original models \underline{underlined}. "E" indicates edit batch size and "AUG" indicates the knowledge augmentation methods. The symbols have the same meaning in the following tables.}
    \label{tab:exp2_1}
\end{table*}

\setlength{\intextsep}{0pt plus 3pt}
\begin{wrapfigure}{r}{0.5\textwidth}
  \centering
  \renewcommand{\arraystretch}{1.2} 
    \scalebox{0.60}{
    \begin{tabular}{cc|cccccc}
    \specialrule{1pt}{0pt}{0pt} 

    \multirow{3}{*}{\centering \textbf{AUG}} & \multirow{3}{*}{\textbf{Method}} & \multicolumn{6}{|c}{\cellcolor{gray!25}\scshape LLaMa-2-7B-chat} \\ \cline{3-8} 
    & & \multicolumn{3}{c}{\textbf{\textsc{MQuAKE}-CF-3K}} & \multicolumn{3}{c}{\textbf{\OurDATA{}}}          \\ \cline{3-8}
    & & 1-E    & 100-E      & All-E  & 1-E    & 100-E  & All-E
    \\ \hline
    \multirow{3}{2cm}{\centering Original}
    & FT       &8.62 &3.34 &0.34      &3.62 &2.05 &0.29          \\
    & ROME      &7.42 &2.14 &0.22   &11.39 &4.64 &0.38    \\ 
    & MEMIT     &\underline{14.33} &\underline{12.17} &\underline{0.59}   &\underline{15.14} &\underline{5.65} &\underline{0.54}  
    \\ \hline
    \multirow{3}{2cm}{\centering\OurMODEL{}}
    & FT      &9.89 &3.62 &0.42    &11.82 &2.66 &0.52          \\
    & ROME      &9.35 &2.14 &0.31   &22.02 &7.02 &0.40    \\ 
    & MEMIT     &\textbf{18.67} &\textbf{15.31} &\textbf{0.77}    &\textbf{29.07} &\textbf{10.19} &\textbf{0.95}  \\
    \bottomrule[1.0pt]
    \end{tabular}
    }
    \caption{Results of parameter-based methods on \textsc{MQuAKE}-CF-3K 
    and~\OurDATA{}. We use multi-hop accuracy (Acc) as 
    the evaluation metric.}
    \label{tab:exp2_2}
\end{wrapfigure}

\noindent {\bf \OurMODEL{} improves knowledge consistency 
for parameter-based methods.}
We also analyze the impact of \OurMODEL{} on the 
parameter-based methods. Corresponding results 
in Figure~\ref{tab:exp2_2} show that
{for \{1,100,All\}-E,~\OurMODEL{} improves 
the performance of the parameter-based methods by 
\{30.29\%, 25.80\%, and 30.50\% \} on \textsc{MQuAKE-CF-3K}; 
\{92.00\%, 80.35\%, and 75.93\% \} on \OurDATA{} respectively.
Its impact is more evident for MEMIT~\cite{meng2022mass}, which yields a stable improvement in performance even under 
a large number of edits. 
\eat{We attribute this performance improvement to the stable editing ability of MEMIT in large editing settings.}
Overall, these results indicate that~\OurMODEL{} is able 
to solve some of the key limitations of parameter-based 
KE methods {on MQA under KE}, highlighted in previous work \cite{zhong2023mquake,gu2023pokemqa}. 

Correlating these results, we observe that for a large 
number of edits, the performance of parameter-based 
systems degrade more compared to that of the 
memory-based methods. This finding is also consistent 
with the previous research~\cite{cheng2024multi}.

\textbf{Back tracking is more challenging than forward tracking.} 
We also analyze the impact of~\OurMODEL{} 
on multiple different scenarios of~\OurDATA{}, 
explained in Section~\ref{sec:benchmark}.
Results in Figure~\ref{fig:bargraph} (a) show that 
compared to the original method, the 
performance improvement incurred by~\OurMODEL{} 
is more prominent for Left/Both-Edit situations
compared to that of No/Right-Edit situations. 
An underlying reason in this regard is the fact that 
the Right-Edit situation requires the process of back tracking, indicating relatively 
lower success rate than that of forward tracking.
To further explore this phenomenon, we examined 
some failing cases with back tracking and found 
that it is primarily caused by many-to-1 relations, 
\eg for the question 
"\emph{Who holds the nationality of America?}" there 
are many people who match this question.

To our surprise, for No-Edit situation (\ie no knowledge 
edits involved), using~\OurMODEL{} leads to slight 
decrease in accuracy.
A probable justification in this regard is the fact that the incorporation of additional knowledge in the edit memory deteriorated the 
retrieval performance, resulting in a decrease in the overall accuracy of the end model.

\textbf{Influence of the correlation threshold.}
The correlation threshold $\delta$ in 
Equation~\eqref{Eq:threshold}, plays a crucial role in 
controlling the final amount and quality of 
additional knowledge generated by~\OurMODEL{}.
Thus, we conduct experiments to explore the 
influence of different values of $\delta$ 
using the variant of~\OurMODEL{} using 
\textsc{LLama-2} and~\OurDATA{}. 
The results in Figure~\ref{fig:bargraph} (b) and (c)
show that~\OurMODEL{} yields the best possible 
performance for value of $\delta$ = 0.8, while 
augmenting the size of edit memory by approximately 3.9$\times$ times.

We also observe that increasing the value of $\delta$ beyond 0.8 will significantly decrease the number of activated rules 
$\Sigma_{e}$ for each edit $e$, 
which in turn limits the amount of newly generated 
knowledge correlated with the edits, thus 
deteriorating the end-performance of~\OurMODEL{}.
On the other hand, decreasing the value of $\delta$
results in an increased number of activated rules 
$\Sigma_{e}$, yielding surplus amount of 
augmented knowledge. This will also limit the 
retrieval accuracy of memory-based methods. 
Overall, higher values of $\delta$ have a more 
detrimental impact on the performance compared to the 
lower ones.

\begin{figure}
    \centering
    \begin{subfigure}{0.287\textwidth}
  \centering
  \includegraphics[width=\linewidth]{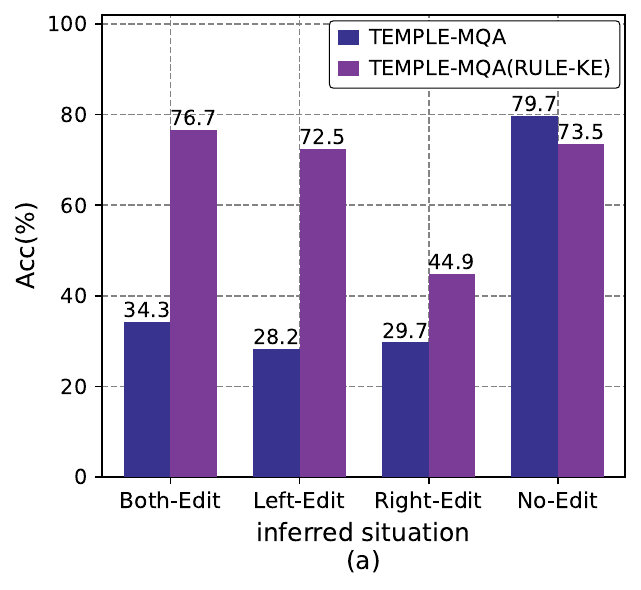}
  
\end{subfigure}%
\begin{subfigure}{0.30\textwidth}
  \centering
  \includegraphics[width=\linewidth]{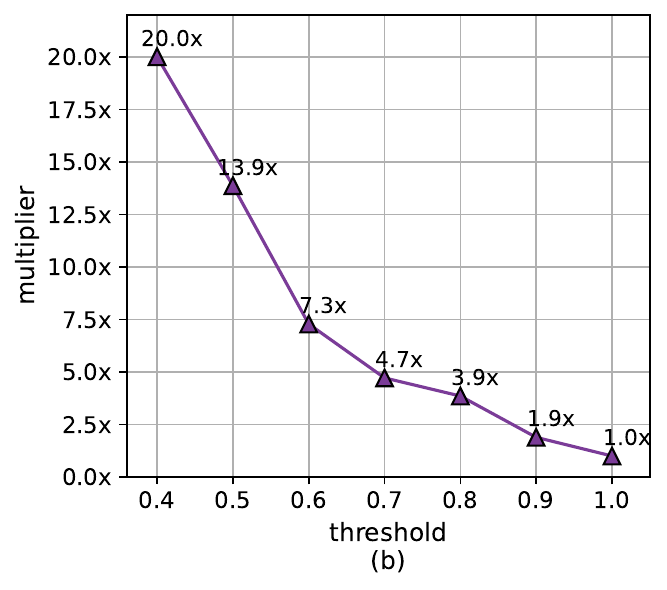}

\end{subfigure}%
\begin{subfigure}{0.28\textwidth}
  \centering
  \includegraphics[width=\linewidth]{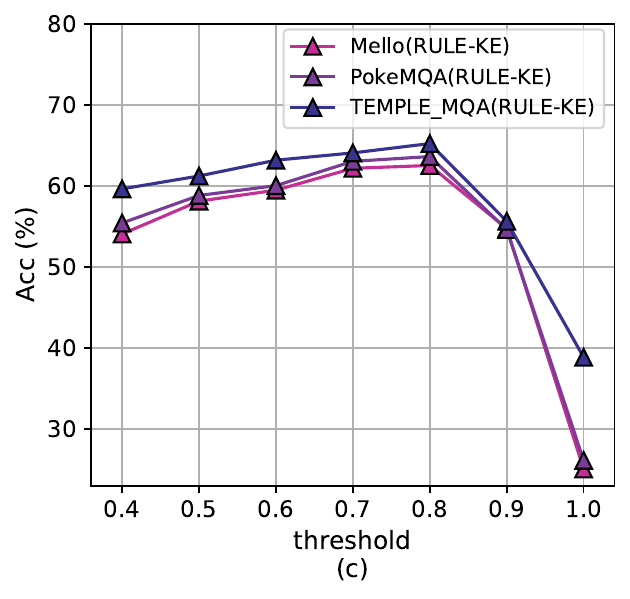}

\end{subfigure}
    \caption{{(a)} Avg., multi-hop accuracy (Acc) for memory-based 
    methods for different inferred situations in~\OurDATA{}.
    {(b)} Increase in size of the edit memory as a multiplier 
    of original size with varying threshold $(\delta)$, for 
    all methods. 
    {(c)} Multi-hop accuracy (Acc) for~\OurMODEL{} using~\OurDATA{} 
    against different values of threshold $(\delta)$. 
    All results are computed using~\textsc{LLama-2-7B-chat}.}
    \label{fig:bargraph}
\end{figure}

\begin{figure}[h]
    \centering
    \begin{subfigure}{0.28\textwidth}
        \centering
        \includegraphics[width=\linewidth]{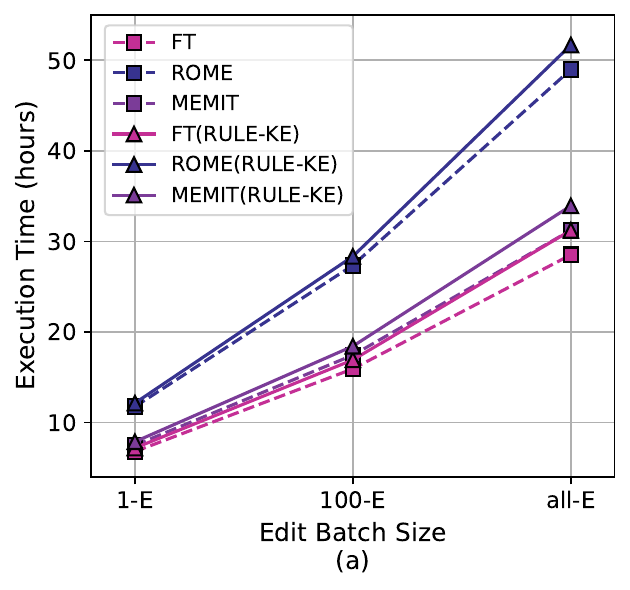}
    \end{subfigure}%
    \begin{subfigure}{0.295\textwidth}
        \centering
        \includegraphics[width=\linewidth]{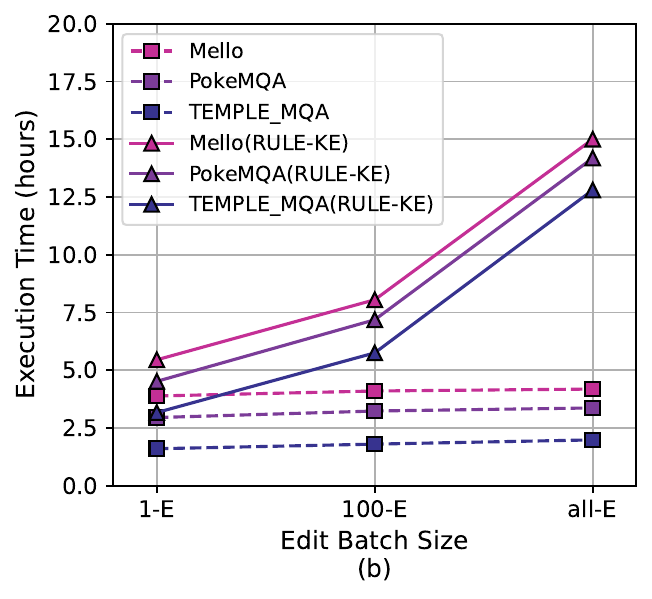}
    \end{subfigure}
    \caption{(a) Execution time of~\OurMODEL{} with parameter-based methods using~\OurDATA{} and~\textsc{LLama-2-7B} under varying 
    edit batch sizes. 
    (b) Execution time overhead of~\OurMODEL{} with memory-based 
    methods using~\OurDATA{} and \textsc{GPT-3.5-turbo-instruct} under varying edit batch sizes.}
    \label{fig:efficiency}
\end{figure}

\noindent{\bf{Efficiency analysis of \OurMODEL{}.}}
Finally, we conduct experiments to explore the efficiency of~\OurMODEL{}. Note that for these results, we ignore the 
time it takes to mine the logic rules, as this process 
is used only once, \ie only at the start of model training.

Figure~\ref{fig:efficiency} (a) shows that for the original variants of parameter-based 
methods, the running time increases with the increase 
in the number of edits. However, the increase in the 
execution time incurred by~\OurMODEL{} is almost 
negligible for edits with different batch sizes.
The results in Figure~\ref{fig:efficiency} (b) show that for the original variants of the memory-based methods, the execution time is almost not impacted by the edit batch size and is primarily
upper-bounded by API calls and/or model inference time. For the model augmented by~\OurMODEL{}, the execution time 
increases by 1.6 folds and 4 folds under 100-E and All-E
settings compared to that of 1-E, respectively.

Overall, these results show that although 
for some cases, the running time of~\OurMODEL{} is 
slightly higher than the original variants, it is still 
within the acceptable range while, at the same time, the 
end model offers significant improvements in the 
performance.


\eat{Overall, these results show that the overall running time of~\OurMODEL{} is withing acceptable range.}
\eat{

it's notable that,~\OurMODEL{} only require 
run one time to generate correlated knowledge, which 
can be used repeatedly for downstream editing methods.

}

\eat{

As we increase the size of the 

As the 
threshold decreases to around 0.4, the amount of additional 
knowledge increases rapidly, but the performance decreases 
slightly. As the threshold increases rapidly to 1.0, the 
performance decreases rapidly.

\di{explain these two observations, why a smaller threshold can be bad? }

}

\eat{
The result in 
Experimental results show that the most suitable threshold 
on the \OurDATA{} is around 0.8, with the highest accuracy, 
and the amount of knowledge added 
is about $3\times$ the original edited knowledge. 

}
\eat{on the activation stage of~\OurMODEL{} in Equation~\eqref{Eq:threshold} is critical 
as it decides the final amounts and quality of 
additional knowledge.}

\eat{exhibit 
interesting findings 
is that memory-based 
methods with~\OurMODEL{} are not as good under 
no/right-edit as under left/both-edit.}

\eat{
\textbf{Without using \OurMODEL{} to enhance model.} We first test the performance of each method on \textsc{MQuAKE} and \OurDATA{} and present the results in Table~\ref{tab:exp1}. The results show that each method performs significantly better on \textsc{MQuAKE} than on \OurDATA{} across all the settings. This is because one of the hops in each multi-hop question of \OurDATA{} is related to derived knowledge. Without rules, it is difficult for the model to obtain derived knowledge according to its existing knowledge and external edited knowledge, leading to wrong answer in the inference process.

\textbf{Enhanced models with \OurMODEL{}.} We then enhance models with \OurMODEL{} and report the results in Table~\ref{tab:exp2_1}. These results clearly show that using \OurMODEL{} significantly enhances the performance of the model, increasing the accuracy by an average of \%, \% and \% under the three settings of 1-edited, 100-edited and All-edited respectively. On \textsc{MQuAKE} dataset, we find that the improvement in accuracy is not significant. This is reasonable because \textsc{MQuAKE} dataset is rule sparse, only few rules are helpful to answer multi-hop questions.

\textbf{Enhanced models with \OurMODEL{} (\textit{w/o rules}).} We finally consider the performance of the model after enhancing with \OurMODEL{} (\textit{w/o rules}), the results are shown in Table ~\ref{tab:exp2_2}. We can see that the performance of each model on \OurMODEL{} is improved, but the improvement is not as great as enhancing with \OurMODEL{}. This is because the knowledge of the model cannot be used to reason with edits without rules and we can only combine  related edits semantically, the single-activation instance cannot be solved well. The accuracy of each method on \textsc{MQuAKE} is also been improved, because combining related edits is helpful to bridge the gap between the coarse grained inference plan and fine grained edits, effectively improving the reliability of the model.
}

\eat{
We also investigate the 
In order to further demonstrate the advantage of \OurMODEL{}, we conduct experiment to contrast effects of \OurMODEL{} and \BaselineMODEL{} to enhance memory-based methods on different scenarios on \OurDATA{}, whose results are shown in figure \ref{fig:bargraph} (a). \OurMODEL{} achieve an improvement of \% and \% on left-activation and right-activation cases respectively, however \BaselineMODEL{} may even degrade the original performance in the this two cases. This phenomenon is because of that rule guide \OurMODEL{} to combine edited knowledge and original knowledge, and for that reason with the enhancement of \OurMODEL{}, the editing methods can handle left-activation and right-activation condition.

\warn{
\textbf{Back tracking is more challenge than forward tracking.} Another interesting findings is that memory-based methods with \OurMODEL{} are not as good under no/right activation as it is under left/both activation. In no-activation condition. There are no edits involved in no-activation, thus, the \OurMODEL{} enhancement leads to a decrease in accuracy due to the incorporation of additional knowledge. The results on Right-activation is not as good as left-activation is because right-activation required additional knowledge on the process of back tracking, which have lower success rate than forward tracking. To explore the reason for that, we examine the failed case when using back tracking and find that it is due to the many-to-1 relation, e.g. "Who hold the the nationality of America?", there are many people match this question.}

\warn{
\textbf{Exploration of the influence of activation rule threshold.} The activation threshold on activation stage of \OurMODEL{} is very important because it decide the final amounts and quality of additional knowledge. Thus, we conduct experiment to explore the final results on this hyper-parameters with varying numbers, whose results are shown in Figure \ref{fig:bargraph} (b) and (c). The experimental results show that the most suitable threshold on the \OurDATA{} is around 0.8, and the amount of knowledge added is about 3 times the original edited knowledge. As the threshold decreases to around 0.4, the amount of additional knowledge increases rapidly and the performance decreases slightly. As the threshold increases rapidly to 1.0, the performance decreases rapidly.}

\eat{
\textbf{Without using \OurMODEL{} to enhance model.} We first test the performance of each method on \textsc{MQuAKE} and \OurDATA{} and present the results in Table~\ref{tab:exp1}. The results show that each method performs significantly better on \textsc{MQuAKE} than on \OurDATA{} across all the settings. This is because one of the hops in each multi-hop question of \OurDATA{} is related to derived knowledge. Without rules, it is difficult for the model to obtain derived knowledge according to its existing knowledge and external edited knowledge, leading to wrong answer in the inference process.

\textbf{Enhanced models with \OurMODEL{}.} We then enhance models with \OurMODEL{} and report the results in Table~\ref{tab:exp2_1}. These results clearly show that using \OurMODEL{} significantly enhances the performance of the model, increasing the accuracy by an average of \%, \% and \% under the three settings of 1-edited, 100-edited and All-edited respectively. On \textsc{MQuAKE} dataset, we find that the improvement in accuracy is not significant. This is reasonable because \textsc{MQuAKE} dataset is rule sparse, only few rules are helpful to answer multi-hop questions.

\textbf{Enhanced models with \OurMODEL{} (\textit{w/o rules}).} We finally consider the performance of the model after enhancing with \OurMODEL{} (\textit{w/o rules}), the results are shown in Table ~\ref{tab:exp2_2}. We can see that the performance of each model on \OurMODEL{} is improved, but the improvement is not as great as enhancing with \OurMODEL{}. This is because the knowledge of the model cannot be used to reason with edits without rules and we can only combine  related edits semantically, the single-activation instance cannot be solved well. The accuracy of each method on \textsc{MQuAKE} is also been improved, because combining related edits is helpful to bridge the gap between the coarse grained inference plan and fine grained edits, effectively improving the reliability of the model.
}
\begin{figure}
    \centering
    \begin{subfigure}{0.287\textwidth}
  \centering
  \includegraphics[width=\linewidth]{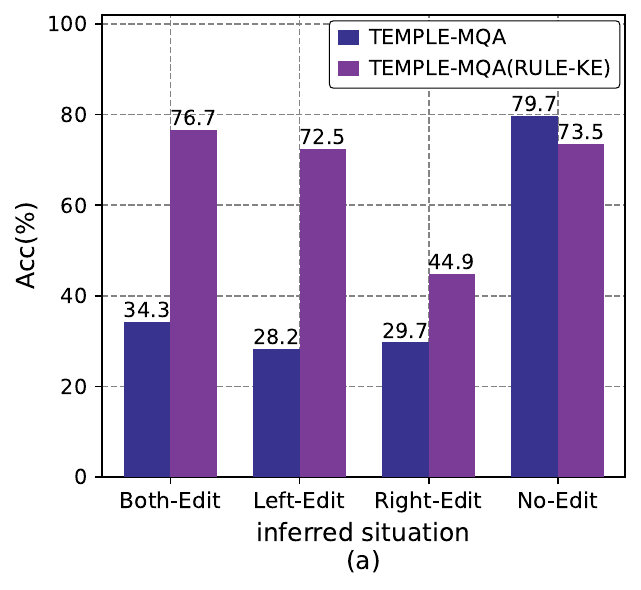}
  
\end{subfigure}%
\begin{subfigure}{0.30\textwidth}
  \centering
  \includegraphics[width=\linewidth]{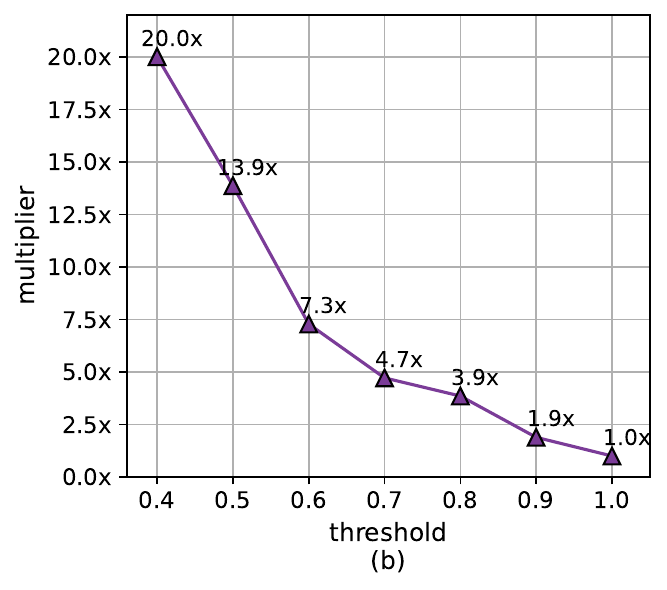}

\end{subfigure}%
\begin{subfigure}{0.28\textwidth}
  \centering
  \includegraphics[width=\linewidth]{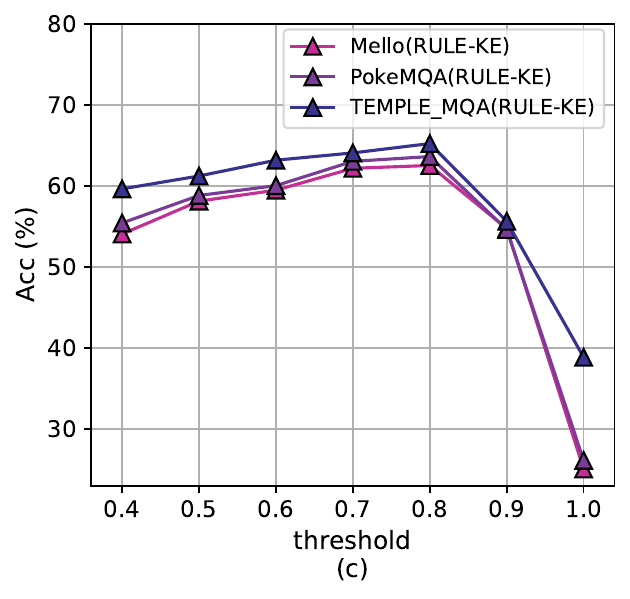}

\end{subfigure}
    \caption{{(a)} Avg., multi-hop accuracy (Acc) for memory-based 
    methods for different inferred situations in~\OurDATA{}.
    {(b)} Increase in size of the edit memory as a multiplier 
    of original size with varying threshold $(\delta)$, for 
    all methods. 
    {(c)} Multi-hop accuracy (Acc) for~\OurMODEL{} using~\OurDATA{} 
    against different values of threshold $(\delta)$. 
    All results are computed using~\textsc{LLama-2-7B-chat}.}
    \label{fig:bargraph}
\end{figure}

\begin{figure}[h]
    \centering
    \begin{subfigure}{0.28\textwidth}
        \centering
        \includegraphics[width=\linewidth]{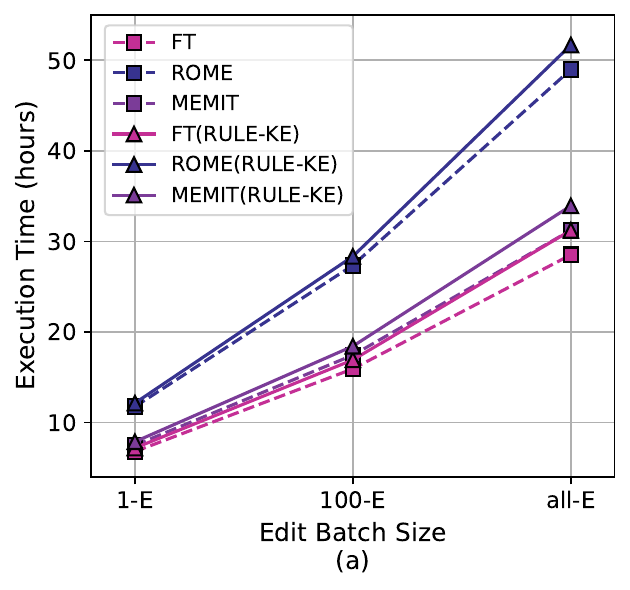}
    \end{subfigure}%
    \begin{subfigure}{0.295\textwidth}
        \centering
        \includegraphics[width=\linewidth]{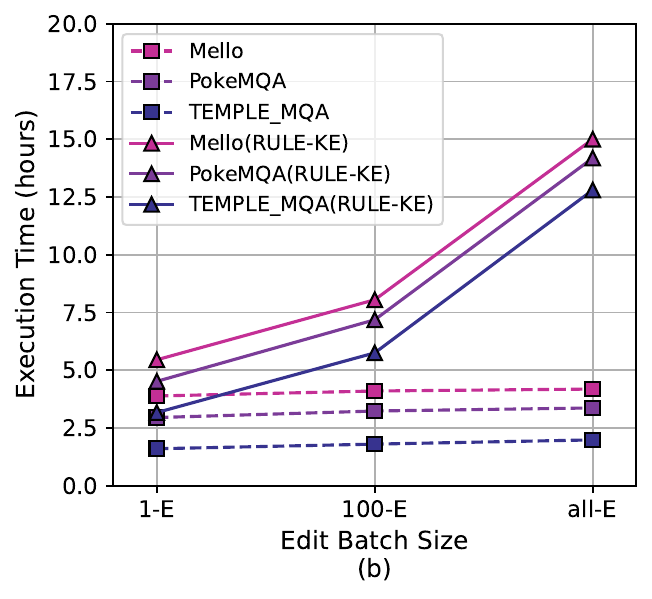}
    \end{subfigure}
    \caption{(a) Execution time of~\OurMODEL{} with parameter-based methods using~\OurDATA{} and~\textsc{LLama-2-7B} under varying 
    edit batch sizes. 
    (b) Execution time overhead of~\OurMODEL{} with memory-based 
    methods using~\OurDATA{} and \textsc{GPT-3.5-turbo-instruct} under varying edit batch sizes.}
    \label{fig:efficiency}
\end{figure}

}

\eat{
By the way, The vast editings situations also lead to huge time-consuming for parameter-based methods, about 24 hours for test only one setting.}

\eat{
\textbf{Parameter-based methods perform poorly.} Comparing to memory-based methods, parameter-based methods exhibit inferior performance across almost all settings. This is consistent with prior research, indicating that parameter-based methods fail to effectively leverage injected knowledge for inference, especially in cases with numerous edits. Despite employing derived knowledge for model editing, parameter-based methods still proves ineffective as the number of edits increases. As shown in Table~\ref{tab:exp2_2}, MEMIT outperforms all parameter-based methods, owing to...}

\eat{which also brings retrieval burden.
this is a trade-off 
The statistics show that the size of edit memory has become 
three times the size on average.}

\eat{the retrieval performance can be improved,~\OurMODEL{} 
can have greater effect. 
For example, }

\eat{In addition, in order to quantify the improvement 
in performance attributable to logic rules, 
we also use a simple variant 
of the proposed model dubbed \BaselineMODEL{} as a baseline model. 
Further details of this model are explained in 
Appendix~\ref{Appendix:model-wo-rules}.}

\eat{Corresponding results 
with enhancement on original, \BaselineMODEL{} and \OurMODEL{}, 
where results for \textsc{MQuAKE} and \OurDATA{} are shown in Table~\ref{tab:exp2_1}.}

\eat{The experimental results of the parameter-based methods and the memory-based methods are presented in Table ~\ref{tab:exp2_2} and Table ~\ref{tab:exp2_1} respectively, which demonstrates that there is a significant improvement for each method using \OurMODEL{}.

Specifically, in \OurDATA{}, using \OurMODEL{} to enhance the parameter-based methods resulted in an average improvement of \%, which is higher than the \% improvement achieved by the baseline method.}
\section{Conclusion}
\label{sec:conclusion}
In this paper, we proposed~\OurMODEL{} that 
leverages logic rules for knowledge consistency 
in order to perform knowledge editing in a 
performance-enhanced fashion.
\OurMODEL{} is a plug-and-play framework that can be 
used with existing MQA methods under KE as 
a \emph{cherry on the top}.
Experimental results using a benchmark and our newly 
curated data set showed that~\OurMODEL{} can augment 
the performance of existing methods on MQA under KE 
by a significant margin.

\bibliographystyle{plain}
\bibliography{reference}

\clearpage
\appendix

\section{Details of Prompts}
\label{sec:prompt}

\definecolor{myTealLight}{rgb}{0.2, 0.8, 0.8} 
\begin{table*}[ht]
    \centering
    \small
    \begin{tabular}{l}
\toprule
\color{red}\textbf{[Demo]}\color{black}\\
Question: $(U.S.,president\ is,\text{<mask>})$, what is the answer for <mask>?\\
Answer: Joe Biden.\\

\color{teal}\textbf{[Instruction]}\color{black}\\
Please refer to the above demo and complete the following task:\\
\color{brown}\textbf{[Task]}\color{black}\\

$(o_{z_m},r_{b_{m+1}},\text{<mask>})$, what is the answer for <mask>?\\

\bottomrule

\end{tabular}
    \caption{{Prompt $\text{P}_{\text{forward}}$ used to determine the next value along the chain in Equation~\ref{eq:forward}}. 
    \color{red}\textbf{[Demo] }\color{black} include several 
    high-quality demonstrations handwritten by humans. $o_{z_m}$ 
    and $r_{b_{m+1}}$ are the inputs for the prompt template.}
    \label{tab:prompt_for_convert_edit}
\end{table*}
\vspace{1em}

\begin{table*}[ht]
    \centering
    \small
    \begin{tabular}{l}
\toprule
\color{red}\textbf{[Demo]}\color{black}\\
Question:  $(\text{<mask>},president\ is,Joe \ Biden)$, what is the answer for <mask>?\\
Answer:  United states.\\

\color{teal}\textbf{[Instruction]}\color{black}\\
Please refer to the above demo and complete the following task:\\
\color{brown}\textbf{[Task]}\color{black}\\

$(\text{<mask>},r_{b_{j}},o_{z_j})$, what is the answer for <mask>?\\

\bottomrule

\end{tabular}
    \caption{{Prompts $\text{P}_\text{back}$ used to determine the previous value along the chain in Equation~\ref{eq:back}}. $r_{b_{j}}$ and $o_{z_j}$ are thes input for prompt template.}
    \label{tab:prompt_for_convert_edit}
\end{table*}

\label{Appendix:oc-ip}

\eat{
\section{Logic Rules}
\label{Appendix:rules}
However, previous work overlooks that one edit not only may affect the reasoning process of $Q$ but also may introduce a new knowledge edit, i.e., an edit $e=r(s,o\rightarrow o^*)$ may lead to a new edit $e^\prime$. For example, $e=\text{Company\_is}(\text{Tom},\text{Amazon}\rightarrow \text{Twitter})$  can lead to $e^\prime=\text{Boss\_is}(\text{Tom},\text{Jeff Bezos}\rightarrow\text{Elon Musk})$. This is because knowledge is not an individual unit, and they are related to each other. Thus, in this paper, we aim to leverage logical rules to represent their association explicitly.}

\section{Retrieval for Forward Tracking and Back Tracking}
\label{Appendix:retrieval}
In this section, we introduce the process of retrieving the edit memory for 
forward tracking and back tracking. For each edit $e\in\mathcal{M}$, we denote $e[s]$, $e[r]$ and $e[o^*]$ as subject, relation and object of edit $e$ respectively.

In the process of forward tracking, we need to whether there is edit $e$ in $\mathcal{M}$, which $e[s]=o_{z_{m}}$ and $e[r]$ match with $r_{b_{m+1}}$. Inspired by previous work \cite{cheng2024multi}, we introduce a two-step retrieval method: 
(i) First, filter out all edits whose subject is not identical to $o_{z_m}$. 
ii) Use dense retrieval to compare the semantic similarity between the relation of the remaining edits and $r_{b_{m+1}}$, then select the most similar one.

\textbf{Step 1: Filtering}. This step extracts a subset $\mathcal{M}_{sub}$ from $\mathcal{M}$ by selecting only the edits that fulfill the subject constraint. 
Formally, this is represented as:
\begin{equation}
   \mathcal{M}_{sub} = \{e \in \mathcal{M}\mid e[s] \in \{o_{z_{m}}\}\cup Alias(o_{z_{m}})  \}
    \label{eq:Re1}
\end{equation}
{where \textit{Alias} represent multiple different alias 
of the entities captured from Wikidata via entity linking~\cite{kolitsas2018end}.}

\textbf{Step 2: Re-ranking}. This step aims to compare the semantic similarity between $e[r]$ and $r_{b_{m+1}}$.
For this, we first encode $r_{b_{m+1}}$ using encoder $E$, as follows:
\begin{equation}
   v_{r}=E(r_{b_{m+1}})
    \label{eq:Ret1}
\end{equation}
Then, we re-rank edits in $\mathcal{M}_{sub}$ based on the cosine similarity ($\text{sim}$) of the relation and select the edit exhibiting highest relational similarity with the retrieved result.
\begin{equation}
 e^*, \eta= \argmax_{e\in \mathcal{M}_{sub}}\  {\text{sim}(E(e[r]), v_r)}, 
    \label{eq:Ret3}
\end{equation}
where $e^*$ is the retrieved edit, and $\eta$ is the 
corresponding similarity compared against a threshold 
$\theta$, as follows:
if $\eta > \theta$, we assume the relevant edit is found, 
and $o_{z_{m+1}}$ is the object of this edit, 
\ie $o_{z_{m+1}} = e^*[o^*]$, otherwise we assume the 
relevant edit is not found.
 
Similar to forward tracking, the objective of back tracking is to obtain $o_{z_{j-1}}$ based on $o_{z_{j}}$ and $r_{b_{j}}$. We first extract subset $\mathcal{M}_{sub}$ to filter out all edits whose object is not identical to $o_{z_{m}}$. Then we match the edits in $\mathcal{M}_{sub}$ according to the semantic similarity between its relation and $r_{b_{j}}$ and select the edit $e^*$ with the highest similarity $\eta$ compared against a threshold $\theta$,
as follows: if $\eta > \theta$, we assume the edit is 
found, \ie $o_{z_{j-1}} = e^*[s]$,
otherwise we assume the relevant edit is not found.

\section{Data Construction of \OurDATA{}}
\label{Appendix:data}
\OurDATA{} is created using Wikidata \footnote{\url{https://www.wikidata.org/wiki/Wikidata:Main_Page}}, 
a large-scale knowledge base encompassing millions of 
real-world entities and their associated knowledge triplets. 
We explain the data curation process as follows:

{\bf Step 1: Collecting high-quality logic rules and relation templates.} We first employ the rule-mining tools to discover a larger number of rules from Wikidata, later manually select a 
small subset of representative rules. 
Table~\ref{tab:rules} illustrated multiple different rules used
for the curation of~\OurDATA{}. 
Following prior work~\cite{zhong2023mquake}, we also collect some common relations from Wikidata and create a question template 
and cloze-style statement template for them, as shown in Table~\ref{tab:templates}.

{\bf Step 2: Constructing relation path template.} Multi-hop questions require a series of reasoning steps to come up with 
the final answer. We refer to all the relations involved in this process as a \textit{relation path template}, \eg for the question: "What is the nationality of the owner of company of Tom?", the corresponding relation path template is "company\_is($z_0$,$z_1$), owner\_is($z_1$,$z_2$), nationality($z_2$,$z_3$)". 
Once given with the start point $z_0$, we can get a potentially valid multi-hop question. 
We get the potential relation path template by arranging all the relation templates obtained in the previous step, and delete 
inappropriate templates, \eg 
"\emph{company$\_$is($z_0$,$z_1$), educated$\_$at$\_$university($z_1$,$z_2$)}" is an inappropriate
template since a company is not a person.

{\bf Step 3: Generating knowledge paths.} Then we use the above relation path template to generate a knowledge path by setting the start point $z_0$ for a specific entity. This is an iterative process which involves determining the specific object $o$ based on the $r(s,?)$, with the subject $s$ and relation $r$ as known. Following the previous work~\cite{cheng2024multi}, we use the database language SPASQL to query the object $o$ for each iterative process. 

{\bf Step 4: Introducing counterfactual edit to the knowledge path.} Next, we use the counterfactual edit to modify some reasoning steps of the knowledge path to mimic the knowledge editing scenario in reality.

{\bf Step 5: Using rules to curate multi-hop questions.} We finally combine some connected relations in the knowledge path into one relation based on logic rules, \eg Combine 
"\emph{father$\_$is(Tom, John), wife$\_$is(John, Amy), company\_is(Amy, Twitter)}" into "\emph{{mother$\_$is}({Tom}, {Amy}), company\_is(Amy, Twitter)}". In this way, 
we can construct a challenging multi-hop question that can 
be used to evaluate the ability of end-models to maintain 
knowledge consistency of KE methods. Finally, we use GPT-4 
to transform them into natural language forms.

\label{Appendix:ourdata_examples}
\begin{table*}[t]
    \centering
    \small
    \scalebox{0.80}{
    \begin{tabular}{cl}
\toprule
\textbf{Question} & Who is the father of the first lady of America?\\
\midrule
\textbf{Edit} & head\_of\_state(America, Joe Biden $\rightarrow$ Albert Einstein)\\
\midrule
\textbf{Rule} & head\_of\_state $\land$ wife\_is $\rightarrow$ first\_lady\_is\\
\midrule
\textbf{Reasoning} & head\_of\_state (America, Albert Einstein) $\land$ wife\_is (Albert Einstein, Elsa Einstein) \\&
$\rightarrow$ first\_lady\_is (America, Elsa Einstein)\\
\midrule
\textbf{Inference Chain} & first\_lady\_is (America, Elsa Einstein)\\& father\_is (Elsa Einstein, Rudolf Einstein) \\
\midrule
\textbf{Answer} & Rudolf Einstein\\
\bottomrule
\end{tabular}}
    \caption{{An example illustration of Left-Edit on~\OurDATA{}.} 
    It shows how~\OurMODEL{} employed logic rules to infer 
    derived knowledge, helpful for the model to generate 
    the correct answer.}
    \label{tab:data_example}
\end{table*}

\section{Additional Experimental Details}
\label{Appendix:add_exp_details}

\subsection{Dataset}
\label{Appendix:dataset}
(i) \textsc{\textsc{MQuAKE}-CF-3K} \cite{zhong2023mquake} is a knowledge editing dataset, which contains a number of k-hop questions ($k\in \{2,3,4\}$) questions, each k-hop question is associated with at least one counterfactual edit. The statistics of \textsc{MQuAKE}-CF-3K are shown in Table~\ref{tab:Statistics_MQuAKE}.\\
(ii) \textsc{\OurDATA{}} is a dataset proposed by us to evaluate the ability of knowledge editing methods to maintain knowledge consistency. Like \textsc{MQuAKE}, \OurDATA{} contains a number of multi-hop questions related to one or more edits. The Statistics of \OurDATA{} are shown in Table~\ref{tab:Statistics_OurDATA}. 

\begin{minipage}[t]{0.5\textwidth}
\centering
\scalebox{0.90}{
\begin{tabular}{cllll}
\toprule[1.0pt]
 \textbf{\#Edits} & \textbf{2-hop} & \textbf{3-hop} & \textbf{4-hop} & \textbf{Total}\\
\hline
 1  &513  &356  &224 &1093 \\
2  &487  &334  &246 &1067 \\  
3  &-  &310  &262 &572 \\
4  &-  &-  &268 &268 \\
\textbf{All}  &1000  &1000  &1000 &3000 \\ 
\bottomrule[1.0pt]
\end{tabular}
}
\captionof{table}{{Statistics of \textsc{MQuAKE}-CF-3K}.}
\label{tab:Statistics_MQuAKE}
\end{minipage}
\begin{minipage}[t]{0.5\textwidth}
\centering
\scalebox{0.90}{
\begin{tabular}{lllll}
\toprule[1.0pt]
 \textbf{Situation} & \textbf{2-hop} & \textbf{3-hop} & \textbf{4-hop} & \textbf{Total}\\
\hline
Left-Edit  &330  &410  &342 &1082 \\
Right-Edit  &357  &290  &244 &891 \\  
Both-Edit  &244  &156  &209 &609 \\
No-Edit  &-  &153  &212 &365 \\
\textbf{All}  &931  &1009  &1007 &2947 \\ 
\bottomrule[1.0pt]
\end{tabular}
}
\captionof{table}{{Statistics of \OurDATA{}}.}
\label{tab:Statistics_OurDATA}
\end{minipage}

\subsection{Large Models}
\label{Appendix:LLMs}
We conduct experiments on the following base language models: (i) \textsc{LLaMa-2-7B} \cite{Touvron2023Llama2O} is a very powerful open source pre-trained large language model, and we use the implementation of huggingface \cite{Wolf2019TransformersSN}. 
(ii) \textsc{LLaMa-2-7B-chat} \cite{Touvron2023Llama2O} is fine-tuned version of \textsc{LLaMa-2-7B} with stronger ability in chat.  
(iii) \textsc{GPT-3.5-turbo-instruct} \cite{Ouyang2022TrainingLM} is a variant of the most capable GPT-3.5 series model, specifically designed to follow instructions and complete tasks thoughtfully.

\subsection{Baseline Methods}
\label{Appendix:baseline}
\noindent{\bf Parameter-based editing method.} 
We take three parameter-based editing methods as baselines.
\begin{enumerate}
    \item {Fine-tuning (FT)} \cite{Zhu2020ModifyingMI} simply perform a gradient descent algorithm on the model based on the new knowledge to update the parameters.
    \item {ROME} \cite{meng2022locating} first locates factual knowledge at a specific layer of the transformer architecture and then updates the feedforward network of this layer to insert new knowledge.
    \item {MEMIT} \cite{meng2022mass} modifies a range of layer feedforward networks and expands ROME to modify a large amount of knowledge.
\end{enumerate}
\noindent{\bf Memory-based editing method.} 
We evaluate the following state-of-the-art 
memory-based editing methods.
\begin{enumerate}
    \item {MeLLo} \cite{zhong2023mquake} uses the plan-and-solve paradigm. When solving a sub-problem, MeLLo first asks the model to generate a candidate answer and then gives the retrieved edits to the model to determine whether they are relevant.
    \item {PokeMQA} \cite{gu2023pokemqa} extend the MeLLo, adopt a two-stage retrieval, and decouple the responsibilities of LLM.
    \item {TEMPLE-MQA} \cite{cheng2024multi} adopts a structural retrieval method, effectively enhancing the accuracy of retrieving relevant edits while addressing the challenge of temporal editing. 
\end{enumerate}

\subsection{Evaluation Metric}
\label{Appendix:evaluation}

\textbf{Multi-hop Accuracy (Acc)} is used to evaluate the performance of the model in multi-hop question-answering. If the answer generated by a model is equal to the target answer ($a^*$) 
or its alias ($Alias(a^*)$), we consider it as a correct answer. 
Assume that the edited model is expressed as $f^*(\cdot)$, 
the multi-hop accuracy is computed as follows:
\begin{equation}
\mathbbm{1}\left[ f^*(q)\in \{a^*\}\cup Alias(a^*) \right ].
\end{equation}
What needs to be emphasized is that each instance in \textsc{MQuAKE}-CF-3K contains three multi-hop questions, and as long as the model can correctly answer one of them, the instance is considered correct. This can be represented by the following formula:
\begin{equation}
\mathbbm{1}\left[\bigvee_{q \in \mathcal{Q}} [f^*(q) \in \{a^*\}\cup Alias(a^*)\right].
\end{equation}
Where $\mathcal{Q}$ represent the multi-hop questions
for each instance.

\begin{table*}[ht]
    \resizebox{0.95\linewidth}{!}{
    \begin{tabular}{l|l|l}
    \toprule 
     Relation & Question template & Cloze-style statement template \\
     \midrule
the\_First\_Lady\_is & Who is the First Lady of [S]? & The First Lady of [S] is \\
head\_of\_state\_is & Who is the current head of state in [S]? & The current head of state in [S] is \\
wife\_is & Who is [S]'s wife? & [S]'s wife is \\
live\_in\_the\_country & Which country does [S] live in? & [S] live in the country of \\
live\_in\_the\_place & Where does [S] live in? & [S] live in \\
located\_in\_the\_country & Which country is [S] located in? & [S] located in the country of \\
studying\_in\_the\_country & Which country is [S] studying in? & [S] study in the country of \\
educated\_at\_the\_university & Which university is [S] educated at? & The university where [S] is educated is \\
native\_language\_is & What is [S]'s native language? & [S]'s countryis \\
holds\_nationality\_in & What is [S]'s nationality? & [S]'s nationality is \\
official\_language\_is & What is the official language of [S]? & The official language of [S] is \\
father\_is & Who is [S]'s father? & [S]'s father is \\
mother\_is & Who is [S]'s mother? & [S]'s mother is \\
husband\_is & Who is [S]'s husband? & [S]'s husband is \\
uncle\_is & Who is [S]'s uncle? & [S]'s uncle is \\
parent\_is & Who is [S]'s parent? & [S]'s parent is \\
brother\_is & Who is [S]'s brother? & [S]'s brother is \\
aunt\_is & Who is [S]'s aunt? & [S]'s aunt is \\
sister\_is & Who is [S]'s sister? & [S]'s sister is \\
grandmother\_is & Who is [S]'s grandmother? & [S]'s grandmother is \\
grandfather\_is & Who is [S]'s grandfather? & [S]'s grandfather is \\
nephew\_is & Who is [S]'s nephew? & [S]'s nephew is \\
sibling\_is & Who is [S]'s sibling? & [S]'s sibling is \\
son\_is & Who is [S]'s son? & [S]'s son is \\
niece\_is & Who is [S]'s niece? & [S]'s niece is \\
daughter\_is & Who is [S]'s daughter? & [S]'s daughter is \\
mother-in-law\_is & Who is [S]'s mother-in-law? & [S]'s mother-in-law is \\
spouse\_is & Who is [S]'s spouse? & [S]'s spouse is \\
father-in-law\_is & Who is [S]'s father-in-law? & [S]'s father-in-law is \\
born\_in\_the\_place & Where is [S]'s birthplace? & [S]'s birthplace is \\
party\_membership\_is & What is [S]'s party membership? & [S]'s party membership is \\
is\_a\_political\_party & Which country is [S] a political party in? & [S] is a political party of \\
located\_in\_the\_continent & Which continent is [S] located in? & [S] located in the continent of \\
located\_in\_the\_continent & Which continent is [S] a country in? & [S] is a country of \\
located\_in\_the\_country & Which country is [S] a university in? & [S] is a university of \\
capital\_is & What is the capital of [S]? & The capital of [S] is \\
affiliated\_with\_the\_religion & Which religion is [S] affiliated with? & [S] is affiliated with the religion of \\
occupation\_is & What is [S]'s occupation? & [S]'s occupation is \\
died\_in\_the\_place & Where is [S]'s place of death? & [S] died in \\
workplace\_is\_at & Where is [S]'s workplace? & [S]'s workplace is at \\
\bottomrule
    \end{tabular}}
    \caption{Question templates and cloze-style statement templates used in \OurDATA{}}
    \label{tab:templates}
\end{table*}

\begin{table}[H]  
\fontsize{8.8}{12}\selectfont
\renewcommand{\arraystretch}{1.2}
\captionsetup{skip=6pt} 
    \begin{tabular}{l}
        \toprule
father\_is($z_0$,$z_1$) $\wedge$ wife\_is($z_1$,$z_2$) $\rightarrow$ mother\_is($z_0$,$z_2$) \\
parent\_is($z_0$,$z_1$) $\wedge$ sister\_is($z_1$,$z_2$) $\rightarrow$ aunt\_is($z_0$,$z_2$) \\
sibling\_is($z_0$,$z_1$) $\wedge$ son\_is($z_1$,$z_2$) $\rightarrow$ nephew\_is($z_0$,$z_2$) \\
parent\_is($z_0$,$z_1$) $\wedge$ brother\_is($z_1$,$z_2$) $\rightarrow$ uncle\_is($z_0$,$z_2$) \\
mother\_is($z_0$,$z_1$) $\wedge$ husband\_is($z_1$,$z_2$) $\rightarrow$ father\_is($z_0$,$z_2$) \\
sibling\_is($z_0$,$z_1$) $\wedge$ daughter\_is($z_1$,$z_2$) $\rightarrow$ niece\_is($z_0$,$z_2$) \\
parent\_is($z_0$,$z_1$) $\wedge$ mother\_is($z_1$,$z_2$) $\rightarrow$ grandmother\_is($z_0$,$z_2$) \\
parent\_is($z_0$,$z_1$) $\wedge$ father\_is($z_1$,$z_2$) $\rightarrow$ grandfather\_is($z_0$,$z_2$) \\
spouse\_is($z_0$,$z_1$) $\wedge$ mother\_is($z_1$,$z_2$) $\rightarrow$ mother-in-law\_is($z_0$,$z_2$) \\
spouse\_is($z_0$,$z_1$) $\wedge$ father\_is($z_1$,$z_2$) $\rightarrow$ father-in-law\_is($z_0$,$z_2$) \\
head\_of\_state\_is($z_0$,$z_1$) $\wedge$ wife\_is($z_1$,$z_2$) $\rightarrow$ the\_First\_Lady\_is($z_0$,$z_2$) \\
holds\_nationality\_in($z_0$,$z_1$) $\wedge$ official\_language\_is($z_1$,$z_2$) $\rightarrow$ native\_language\_is($z_0$,$z_2$) \\
party\_membership\_is($z_0$,$z_1$) $\wedge$ is\_a\_political\_party($z_1$,$z_2$) $\rightarrow$ holds\_nationality\_in($z_0$,$z_2$) \\
live\_in\_the\_place($z_0$,$z_1$) $\wedge$ located\_in\_the\_country($z_1$,$z_2$) $\rightarrow$ live\_in\_the\_country($z_0$,$z_2$) \\
born\_in\_the\_place($z_0$,$z_1$) $\wedge$ located\_in\_the\_country($z_1$,$z_2$) $\rightarrow$ holds\_nationality\_in($z_0$,$z_2$) \\
located\_in\_the\_country($z_0$,$z_1$) $\wedge$ located\_in\_the\_continent($z_1$,$z_2$) $\rightarrow$ located\_in\_the\_continent($z_0$,$z_2$) \\
educated\_at\_the\_university($z_0$,$z_1$) $\wedge$ located\_in\_the\_country($z_1$,$z_2$) $\rightarrow$ studying\_in\_the\_country($z_0$,$z_2$) \\
\bottomrule

    \end{tabular}
    \caption{Logic rules in~\OurDATA{}.}
    \label{tab:rules}
\end{table}
\section{Limitations}
\label{limit}
While large language models (LLMs) have garnered widespread 
attention due to their exceptional knowledge comprehension capabilities, research shows they experience hallucinations 
and may generate incorrect knowledge.
While~\OurMODEL{} is an attempt to 
overcome this limitation, 
yet employing our model may not completely eradicate 
this problem.
This challenge stems from: 
(i) inherent limitations underlying LLMs modeling assumptions,
(ii) in-consistency in the data used for training the large models,
however, these problems could be alleviated as the data and
capabilities of LLMs continue to improve in the future.

Besides, although Large Language Models (LLMs) have garnered widespread attention for their remarkable capacity for knowledge comprehension \cite{yang2024human,yang2024moral,cheng2024multi,ali2024prompt,yang2024dialectical}, enabling tailored solutions across various applications, they also face critical issues such as privacy concerns \cite{hu2023differentially}, and explainability \cite{hu2023seat,lai2023faithful,hu2023improving}. LLM applications typically involve data containing sensitive information, necessitating effective solutions to safeguard privacy \cite{xu2023llm}. One promising approach to address this challenge is the design of Differentially Private (DP) algorithms \cite{dwork2006calibrating}. DP offers provable protection against identification and is resilient to arbitrary auxiliary information that may be available to attackers. While there have been numerous studies on DP machine learning \cite{hu2022high,wang2020differentially,wang2021estimating,su2022faster,hu2023privacy} and DP deep learning \cite{xiang2024does,xiang2023practical}, most of these efforts have primarily focused on either continuous tabular data or image data. Unfortunately, less attention has been given to adapting variants of DP algorithms to the context of Natural Language Processing (NLP) and the text domain. Addressing this gap is crucial as text data presents its own unique challenges and characteristics that necessitate specialized privacy-preserving techniques. By developing and refining DP algorithms tailored to NLP tasks, we can enhance the privacy protections of LLMs and enable their responsible and ethical deployment across various domains. We will leave it for future work.
\clearpage

\end{document}